\documentclass[a4paper,fleqn]{cas-dc}
\renewcommand{\printorcid}{}

\usepackage[numbers]{natbib}

\def\tsc#1{\csdef{#1}{\textsc{\lowercase{#1}}\xspace}}
\tsc{WGM}
\tsc{QE}

\usepackage{caption}
\usepackage{subcaption}

\begin{document}
\let\WriteBookmarks\relax
\def\floatpagepagefraction{1}
\def\textpagefraction{.001}

\shorttitle{DeepFH Segmentations for Superpixel-based Object Proposal Refinement}    

\shortauthors{Wilms and Frintrop}  

\title [mode = title]{DeepFH Segmentations for Superpixel-based Object Proposal Refinement}  



%

\author[1]{Christian Wilms}

\ead{wilms@informatik.uni-hamburg.de}




\author[1]{Simone Frintrop}

\ead{frintrop@informatik.uni-hamburg.de}


\affiliation[1]{organization={University of Hamburg, Department of Informatics},
            addressline={Vogt-Koelln-Str. 30}, 
            city={Hamburg},
            postcode={22527}, 
            country={Germany}}

\cortext[1]{Corresponding author}

\begin{abstract}
Class-agnostic object proposal generation is an important first step in many object detection pipelines. However, object proposals of modern systems are rather inaccurate in terms of segmentation and only roughly adhere to object boundaries. Since typical refinement steps are usually not applicable to thousands of proposals, we propose a superpixel-based refinement system for object proposal generation systems. Utilizing precise superpixels and superpixel pooling on deep features, we refine initial coarse proposals in an end-to-end learned system. Furthermore, we propose a novel DeepFH segmentation, which enriches the classic Felzenszwalb and Huttenlocher (FH) segmentation with deep features leading to improved segmentation results and better object proposal refinements. On the COCO dataset with LVIS annotations, we show that our refinement based on DeepFH superpixels outperforms state-of-the-art methods and leads to more precise object proposals.
\end{abstract}


\maketitle

\section{Introduction}
\label{sec:intro}
\begin{figure}[t]
\centering
\begin{subfigure}{0.23\textwidth}
\centering
\includegraphics[width=\textwidth]{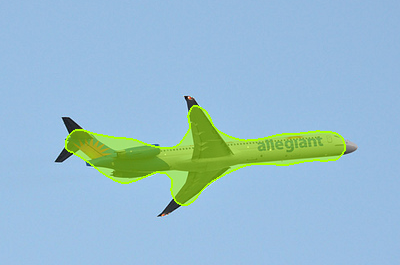}
\caption{without our refinement}
\label{fig:comparePx}
\end{subfigure}
\begin{subfigure}{0.23\textwidth}
\centering
\includegraphics[width=\textwidth]{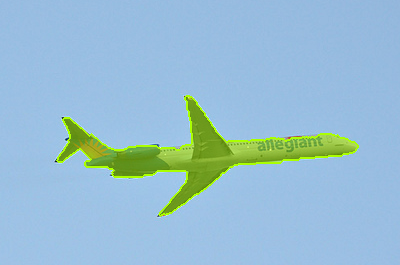}
\caption{with our refinement}
\label{fig:comapreSpx}
\end{subfigure}
\caption{AttentionMask object proposal without (a) and with (b) our superpixel-based refinement. Note the precise segmentation of fine details after applying our refinement.}
\label{fig:comparePxSpx}
\end{figure}

Object proposal generation serves as the basis for many object detection or instance segmentation systems~\cite{girshick2015fast,ren2015faster,he2017mask,chen2018masklab,WilmsAirlinesICPR2020}. The task describes the class-agnostic generation of possible object locations, which reduce the search space for subsequent methods. Since the task is class-agnostic, different to, e.g., instance segmentation, no class information is utilized. This leads to better generalization to unseen classes~\cite{Osep19ICRA}. Since the emergence of deep learning, a number of CNN-based methods for object proposal generation were proposed~\cite{Pinheiro2015-deepmask,Pinheiro2016-sharpmask,Hu2017-fastmask,WilmsFrintropACCV2018,gidaris2016attendbmvc,li2017zoom}. However, most of these systems suffer from imprecise segmentations, especially systems that propose segmentation masks as visible in Fig.~\ref{fig:comparePx}.

A major reason for the imprecise segmentations is the CNN, which leads to a low resolution in the system's segmentation stage. For instance, in~\cite{WilmsFrintropACCV2018} segmentations are based on $10 \times 10$ feature maps, independent of the object's size. However, those low resolution feature maps are semantically rich and important for the systems' overall success. Thus, systems suffer from coarse and imprecise segmentations, despite generating state-of-the-art overall results.

In tasks like semantic segmentation or salient object detection, this problem is tackled by utilizing encoder-decoder architectures~\cite{li2016deep,chen2017deeplab,li2018referring} or CRFs~\cite{hou2017deeply,lin2017refinenet}. However, this is computationally expensive if applied to 1000 proposals per image. Other lines of work include dilated convolutions~\cite{chen2017deeplab,yu2015multi} or iterative refinement \cite{zhang2019canet,kirillov2019pointrend}. Most of these approaches are computationally expensive as well if applied to all proposals or many layers of a network. Thus, a refinement system which precisely captures the object contours and shares computations between the proposals is necessary.

Superpixels, introduced by~\cite{ren2003learning}, capture fine details of objects and serve as an abstraction of the pixels by reducing the number of entities. Furthermore, superpixels can be shared between the proposals of an image, since they only rely on the image content. However, applying superpixels in CNNs is not straightforward, since they lack a lattice structure. However, \cite{he2017std2p,kwak2017weakly,park2017superpixel}~proposed semantic segmentation approaches with information aggregation across superpixels in CNNs.

\begin{figure}[t]
\centering
\includegraphics[width=0.35\textwidth]{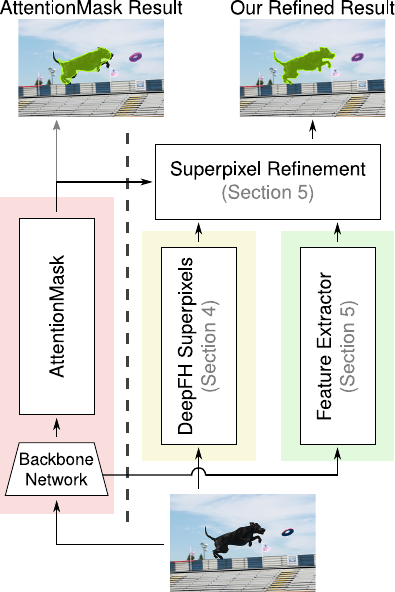}
\caption{Building blocks of our proposed system. We refine object proposals with deep features and superpixels generated with our proposed DeepFH segmentation (cf. Sec.~\ref{sec:meth_deepfh_FH}). The streams are combined in our superpixel refinement module (cf. Sec.~\ref{sec:meth_refinement}) utilizing superpixel pooling and a superpixel classifier leading to precise proposals. The images on the top show the original proposal (left) and our refined version (right).}
\label{fig:abstractSysFig}
\end{figure}

In this paper, we follow this aggregation approach and propose a superpixel-based refinement for object proposals. We start from the coarse proposals of the state-of-the-art object proposal generation system AttentionMask~\cite{WilmsFrintropACCV2018} and add two streams to the system as visualized in Fig.~\ref{fig:abstractSysFig}. The first stream generates precise superpixel segmentations on input image resolution, while the second stream extracts features from the backbone network. In our proposed superpixel refinement module, for each proposal, we aggregate the information from the coarse AttentionMask proposal and the extracted features per superpixel utilizing our superpixel pooling module following~\cite{he2017std2p,kwak2017weakly,park2017superpixel}. For each proposal, this leads to a feature vector per superpixel, which is classified as background or object part with our new superpixel classifier. The evaluation shows superior overall results compared to state-of-the-art systems. Furthermore, the masks adhere better to the object boundaries (see Fig.~\ref{fig:comparePxSpx}), which is outlined by our results on the COCO dataset~\cite{lin2014microsoft} using LVIS annotations~\cite{gupta2019lvis}.

A typical problem when utilizing superpixels for object proposal refinement are compact superpixels. They lead to artificially splitting up uniform areas and increases the number of superpixels without significant gain in segmentation accuracy. Furthermore, superpixels, which cover large areas, have a larger support for information aggregation. Few approaches~\cite{felzenszwalb2004efficient,luengo2016smurfs,WilmsFrintropGCPR2017} have been proposed for generating non-compact superpixels, since it is difficult to control the number of superpixels for evaluation~\cite{stutz2018superpixels}.

As a second novelty, we propose the novel DeepFH segmentation. DeepFH extends the segmentation algorithm by Felzenszwalb and Huttenlocher (FH)~\cite{felzenszwalb2004efficient}, which relies on RGB features for pixel similarity, with semantically richer deep features. We generate deep features with a lightweight encoder-decoder architecture based on a variation of ResNet~\cite{he2016-resnet}. In our evaluation, we show that DeepFH outperforms the original FH and further improves the downstream superpixel-based object proposal refinement. Since FH's general structure is unchanged, FH's properties including non-compact superpixels stay unchanged. 

Our main contributions are a new superpixel-based refinement system for object proposals with precise proposal masks outperforming state-of-the-art and a new deep learning-based segmentation algorithm to support the refinement system. Overall, this paper presents an extended version of our conference paper~\cite{WilmsFrintropICPR2020}. We extend that paper in three major ways:
\begin{itemize}
\item Utilizing deep features with a tailored encode-decoder architecture, leading to novel DeepFH segmentations based on classic FH.
\item An advanced superpixel-based refinement approach for object proposal generation systems combining DeepFH and superpixel pooling for improved results and better boundary adherence.
\item A more detailed evaluation of our overall refinement system w.r.t. the superpixel segmentations.
\end{itemize}

\section{Related Work}
\label{sec:relWork}
This section reviews related approaches for object proposal generation and superpixel segmentation.

\subsection{Object Proposal Generation}
\label{sec:relWork_proposals}
Class-agnostic object proposal generation systems produce bounding boxes~\cite{gidaris2016attendbmvc,li2017zoom} or pixel-precise segmentation masks~\cite{Pinheiro2015-deepmask,Pinheiro2016-sharpmask,Hu2017-fastmask,WilmsFrintropACCV2018} alongside an objectness score per proposal. \cite{Hosang2015PAMI}~give an overview of approaches not utilizing CNNs. Since we propose a system with pixel-precise masks, we focus on these approaches here.

Pre CNN systems usually grouped superpixels based on low-level or mid-level image cues. For instance, \cite{pont2017multiscale}~propose a hierarchical image segmentation with subsequent merging and boundary classification.

Utilizing CNNs, \cite{Pinheiro2015-deepmask}~propose DeepMask, a model with individual heads for inferring objectness and the segmentation mask on top of a backbone network. For generating multiple proposals per image, DeepMask runs on many crops from an image pyramid (multi-shot approach). Due to subsampling in the backbone, the masks are generated on a coarse resolution, leading to imprecise proposals. SharpMask~\cite{Pinheiro2016-sharpmask} tackles this problem by adding a decoder path to refine the segmentation masks. Still, SharpMask utilizes the multi-shot approach.

Feeding overlapping image crops through the backbone is redundant. To increase efficiency, \cite{Hu2017-fastmask}~propose FastMask with a feature pyramid inside the CNN (one-shot approach). \cite{Hu2017-fastmask}~generate the feature pyramid based on the backbone's final layer utilizing pooling. All possible fixed sized windows are extracted from the pyramid to capture objects of different size. For segmentation and objectness scoring, \cite{Hu2017-fastmask}~propose individual heads. Due to the pyramid on top of the backbone, the feature maps for segmentation are coarse, resulting in imprecise masks.

Extending FastMask, \cite{WilmsFrintropACCV2018}~propose AttentionMask. AttentionMask improves the efficiency with attention at each pyramid level. As a result, only relevant windows are  extracted, omitting background areas. Utilizing the freed resources, \cite{WilmsFrintropACCV2018}~add a new level to the pyramid for capturing small objects. The segmentation masks are still generated on $40 \times 40$ maps, independent of the object's size. Thus, despite improved results and efficiency, the masks are still coarse.

Complementing those approaches, we propose a super\-pixel-based refinement to enhance the quality of coarse segmentation masks.

\subsection{Superpixel Segmentation}
\label{sec:relWork_superpixels}
Since the seminal work of~\cite{ren2003learning}, many superpixel segmentation approaches were proposed~\cite{achanta2012slic,van2012seeds,yao2015real,liu2011entropy,felzenszwalb2004efficient,tu2018learning,jampani2018superpixel,yang2020superpixel}. \cite{stutz2018superpixels}~give an overview of approaches not utilizing CNNs. Many superpixel algorithms start from an initial segmentation, which is refined to fit the image content, leading to compact superpixels of uniform size. \cite{achanta2012slic}~start from a uniform grid and apply a local k-means clustering to the pixels. \cite{van2012seeds}~also start from an initial grid and move pixels or pixel blocks between superpixels. Similarly, \cite{yao2015real}~propose an MRF approach with a topology preserving energy term.

In contrast, \cite{liu2011entropy} and~\cite{felzenszwalb2004efficient} merge individual pixels to generate a segmentation. \cite{liu2011entropy}~merge pixels based on the entropy of random walks to generate superpixels of roughly uniform size. The FH algorithm~\cite{felzenszwalb2004efficient} merges pixels based on color or intensity differences. The termination depends on internal and external dissimilarities of pixel clusters. Thus, the superpixel size is not regularized, leading to non-compact superpixels of varying size. \cite{WilmsFrintropGCPR2017}~propose an approach to get a similar effect with other segmentation algorithms.

Few approaches generate superpixels utilizing CNNs. \cite{tu2018learning}~compute pixel affinities and apply~\cite{liu2011entropy} on those affinities. \cite{jampani2018superpixel}~propose a differentiable approach of~\cite{achanta2012slic} with CNN features. \cite{yang2020superpixel}~learn associations between each pixel and superpixels of a uniform grid to infer superpixels.

Different to those approaches, we utilize deep features to enhance the FH segmentation. Thus, the segmentation allows non-compact superpixels and utilizes deep features, which improves our downstream application.

\section{System Overview}
\label{sec:sysOverview}
Our overall system consists of four building blocks as visualized in Fig.~\ref{fig:abstractSysFig}. The first part is the  state-of-the-art object proposal system AttentionMask~\cite{WilmsFrintropACCV2018} (red area in Fig.~\ref{fig:abstractSysFig}), which we use to generate coarse object proposals. Second, we propose a novel DeepFH superpixel segmentation, described in Sec.~\ref{sec:meth_deepfh}, based on deep features and the classic FH algorithm (yellow area  Fig.~\ref{fig:abstractSysFig}). Third, we extract features (see Sec.~\ref{sec:meth_refinement}) from AttentionMask's backbone (green area  Fig.~\ref{fig:abstractSysFig}). Finally, we combine these streams in our novel superpixel-based refinement module described in Sec.~\ref{sec:meth_refinement} (top area in Fig.~\ref{fig:abstractSysFig}). The refinement module takes AttentionMask's coarse proposals and refines them using the deep features and the new DeepFH superpixels utilizing superpixel pooling and our novel superpixel classifier. The output of the overall system are refined object proposals, which better adhere to object boundaries.

\section{DeepFH Segmentation}
\label{sec:meth_deepfh}
\begin{figure}[t]
\centering
\includegraphics[width=0.4\textwidth]{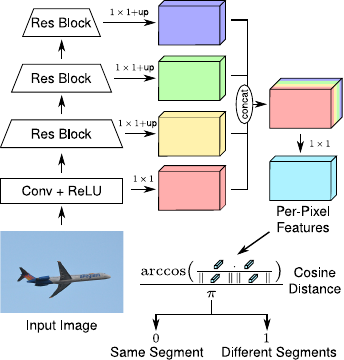}
\caption{Feature extractor for our DeepFH segmentation. The encoder consists of ResNet-18 blocks ($2$ convolution layers with skip connection). After applying a $1 \times 1$ convolution per stage, the features of different stages (different colors) are concatenated and the final features are extracted. Per pixel, these features are used for pixel affinity calculation based on cosine distance.}
\label{fig:sysFigDeepFH}
\end{figure}

This section introduces our novel DeepFH segmentation as a variation of the segmentation algorithm by Felzenszwalb and Huttenlocher (FH)~\cite{felzenszwalb2004efficient} based on CNN-features. For generating a DeepFH segmentation, we run an image through a lightweight encoder-decoder (cf. Fig.~\ref{fig:sysFigDeepFH}). This results in semantically rich per-pixel features at the decoder's output layer (cyan box in Fig.~\ref{fig:sysFigDeepFH}) with input image resolution. We use these features to enrich the color distance between pixels in the classic FH algorithm with the cosine distance in the deep feature space for a combined distance. The rest of the FH algorithm (see Sec.~\ref{sec:meth_deepfh_FH}) is unchanged to keep the algorithm's properties like non-compact superpixels.

\subsection{Feature Extraction}
\label{sec:meth_deepfh_featExtraction}
For extracting semantically rich CNN-based features, we follow the works on pixel affinity by~\cite{ahn2018learning,ahn2019weakly} for semantic segmentation as well as the approaches of~\cite{tu2018learning,jampani2018superpixel,yang2020superpixel} on CNN-based superpixel segmentation. We propose a simple encoder-decoder with a 4-stage ResNet-18-based backbone for feature extraction as visible on the left in Fig.~\ref{fig:sysFigDeepFH}.  In contrast to ResNet-18, we use only one residual block per stage in the encoder. This is in line with \cite{tu2018learning,jampani2018superpixel,yang2020superpixel}, which propose shallow networks for feature extraction. Furthermore, this architecture generates better results compared to networks with more layers like ResNet-18 or ResNet-34 (see top part of Tab.~\ref{tab:encoderDecoder}).

The decoder part (right part in Fig.~\ref{fig:sysFigDeepFH}) consists of a per-stage feature extraction with a bilinear upsampling, a concatenation across all stages and the final feature extraction layer leading to a 128 dimensional feature vector per pixel. This is similar to the decoder architectures in pixel affinity \cite{ahn2018learning,ahn2019weakly,tu2018learning} and is more powerful than the step-wise integration of feature maps in~\cite{yang2020superpixel} (see bottom part of Tab.~\ref{tab:encoderDecoder}). 

\begin{table}[t]
\renewcommand{\arraystretch}{1.2}
\centering
\caption{Architectures for our feature extraction network's encoder (top) and decoder (bottom), evaluated w.r.t. the downstream object proposal refinement task. AR@$n$ denotes the average recall for the first $n$ proposals.
}
\label{tab:encoderDecoder}
{
\begin{tabular}{@{}lcccccc@{}}
\hline
Architecture &  AR@10$\uparrow$ & AR@100$\uparrow$  & AR@1000$\uparrow$ \\ \hline
Ours & \textbf{0.104} &  \textbf{0.223} & \textbf{0.330} \\
ResNet-18 & 0.102 &  0.222 & 0.324  \\
ResNet-34 & 0.102 &  0.223 & 0.325  \\
\hline
Step-wise~\cite{yang2020superpixel} & 0.102 &  0.221 & 0.323  \\ 
\hline
\end{tabular}
}
\end{table}

On top of the final layer, the cosine distance between feature vectors of adjacent pairs of pixels ($\mathbf{f}_i, \mathbf{f}_j$) is calculated:
\begin{equation}
\delta_{\cos}(\mathbf{f}_i, \mathbf{f}_j) = \frac{1}{\pi}\arccos\left(\frac{\mathbf{f}_i \cdot \mathbf{f}_j}{\|\mathbf{f}_i\| \|\mathbf{f}_j\|}\right).
\end{equation}
The result of the cosine distance is the classification result for the pair of adjacent pixels. If the distance is low, the  pixels belong to the same segment and vice versa. Applying the cosine distance is different to~\cite{ahn2018learning,ahn2019weakly}, which use Euclidean distance. However, since the feature vectors are high-dimensional, cosine distance is favorable (cf. Tab.~\ref{tab:distance}).

\begin{table}[t]
\renewcommand{\arraystretch}{1.2}
\centering
\caption{Cosine distance and Euclidean distance as deep feature distance in DeepFH evaluated w.r.t. the downstream object proposal refinement task. AR@$n$ denotes the average recall for the first $n$ proposals.
}
\label{tab:distance}
{
\begin{tabular}{@{}lcccccc@{}}
\hline
Distance &  AR@10$\uparrow$ & AR@100$\uparrow$  & AR@1000$\uparrow$ \\ \hline
Cosine & \textbf{0.104} &  \textbf{0.223} & \textbf{0.330} \\
Euclidean & 0.102 &  0.218 & 0.320  \\
\hline
\end{tabular}
}
\end{table}

\subsection{Extension of FH Algorithm}
\label{sec:meth_deepfh_FH}
The original FH algorithm~\cite{felzenszwalb2004efficient} is a graph-based segmentation method, in which adjacent pixels or pixel clusters are greedily merged. As criterion for merging, simple color or intensity difference is used. For RGB images, the difference between the pixels $\mathbf{p}_i$ and $\mathbf{p}_j$ representing the RGB values as vector is calculated as
\begin{equation}
\delta_{\text{FH}} = \| \mathbf{p}_i - \mathbf{p}_j \|_2.
\end{equation}
During merging, the largest dissimilarity within a cluster of pixels and the smallest dissimilarity between two adjacent clusters of pixels are compared. If a certain threshold-based criterion is met, the clusters are merged. The algorithm terminates once no pair of clusters meets the criterion. Since all steps are based on the initial pixel distance, by only enhancing this distance calculation, the outcome of the FH algorithm can be improved.

Enhancing the distance calculation for our DeepFH, we utilize the CNN-features $\mathbf{f}_i,$ and $\mathbf{f}_j$ for adjacent pixels $\mathbf{p}_i$ and $\mathbf{p}_j$ extracted from the encoder-decoder (see Sec.~\ref{sec:meth_deepfh_featExtraction}). The features are extracted from the final feature map (cyan box in Fig.~\ref{fig:sysFigDeepFH}). Since the features were learned based on cosine distance, we add the cosine distance between pixels' features as a second term to the original distance $\delta_{\text{FH}}$:
\begin{equation}
\label{eq:newFH}
\delta_{\text{DeepFH}} = (1-\alpha) \| \mathbf{p}_i - \mathbf{p}_j \|_2 + \alpha \delta_{\cos}(\mathbf{f}_i, \mathbf{f}_j).
\end{equation}

This is convenient, since both distances are in the interval $[0,1]$. Thus, the parameter $\alpha$ only controls the influence of the distances. We set $\alpha=0.2$ for best results on the downstream object proposal refinement task. The rest of the FH algorithm stays unchanged. Thus, the runtime is still $\mathcal{O}(n\log n)$ and the properties regarding the segmentation quality proven in~\cite{felzenszwalb2004efficient} are kept.

\subsection{Training}
\label{sec:meth_deepfh_training}
For training the feature extractor, we use binary crossentropy loss to assess the feature vectors' cosine distance between adjacent pixels. For optimization, we use Adam with an initial learning rate of $0.01$, $\beta_1=0.9$, and $\beta_2=0.999$. Similar to~\cite{tu2018learning,jampani2018superpixel,yang2020superpixel}, our network is trained from scratch without pre-trained weights, since pre-training did not improve the results.

The overall goal of our system is to generate better object proposals. Therefore, we did not train on typical segmentation datasets, but generated a segmentation dataset from the COCO dataset~\cite{lin2014microsoft} training images with LVIS annotations~\cite{gupta2019lvis}. The LVIS annotations are significantly more precise, which is important in our pixel affinity-like setup, and contain one binary segmentation per annotated object. As ground truth, we combine all binary segmentations per image, generating an oversegmentation of the image retaining all object boundaries. We train the network on this dataset for 8 epochs. The integration of DeepFH segmentations into the superpixel-based proposal refinement is described in Sec.~\ref{sec:meth_refinement_combination}.

\section{Superpixel-based Object Proposals}
\label{sec:meth_refinement}
\begin{figure}[t]
\centering
\includegraphics[width=0.48\textwidth]{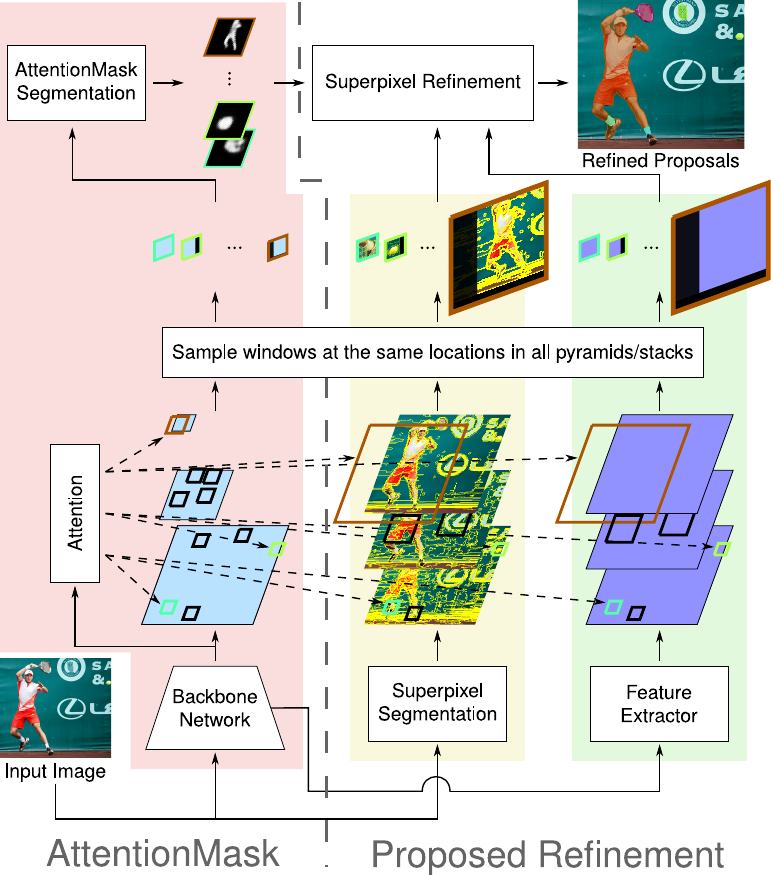}
\caption{Overview of our proposed superpixel-based refinement system. The general structure follows AttentionMask (red area) and is extended by our DeepFH superpixel segmentations (yellow area), our feature extraction (green area) as well as our superpixel refinement module. The system's three streams share the same pyramid/stack structure and are connected by AttentionMask's attention module. The module selects interesting areas in the pyramids/stacks, which are cropped from all streams (colored windows). However, indicated by the different sizes, the windows cropped from the segmentation and feature extraction streams have a higher resolution, leading to more detailed results. Finally, the cropped windows from the three streams are utilized in our superpixel refinement module (see Fig.~\ref{fig:ablaufSpxRefinement}).}
\label{fig:sysFigSpxRefinement}
\end{figure}

This section introduces our superpixel-based refinement for object proposal generation. The refinement connects the object proposal generation system AttentionMask~\cite{WilmsFrintropACCV2018} (see Sec.~\ref{sec:relWork_proposals}) and the DeepFH segmentation described in Sec.~\ref{sec:meth_deepfh_FH}. It starts from the coarse proposals of AttentionMask, which are $40 \times 40$ windows per proposal. AttentionMask is visualized in the red part of Fig.~\ref{fig:sysFigSpxRefinement}, which depicts our overall refinement system. The $40 \times 40$ windows contain per-pixel probabilities for being part of an object and represent only a certain area of the input image. In parallel, we generate precise DeepFH superpixel segmentations for each level of AttentionMask's feature pyramid with different resolutions, capturing fine details of the input image (yellow part in Fig.~\ref{fig:sysFigSpxRefinement}). In a third stream, displayed in the green part of Fig.~\ref{fig:sysFigSpxRefinement}, features are extracted from AttentionMask's backbone for each level of the feature pyramid to support the refinement.

\begin{figure*}[t]
\centering
\includegraphics[width=0.8\textwidth]{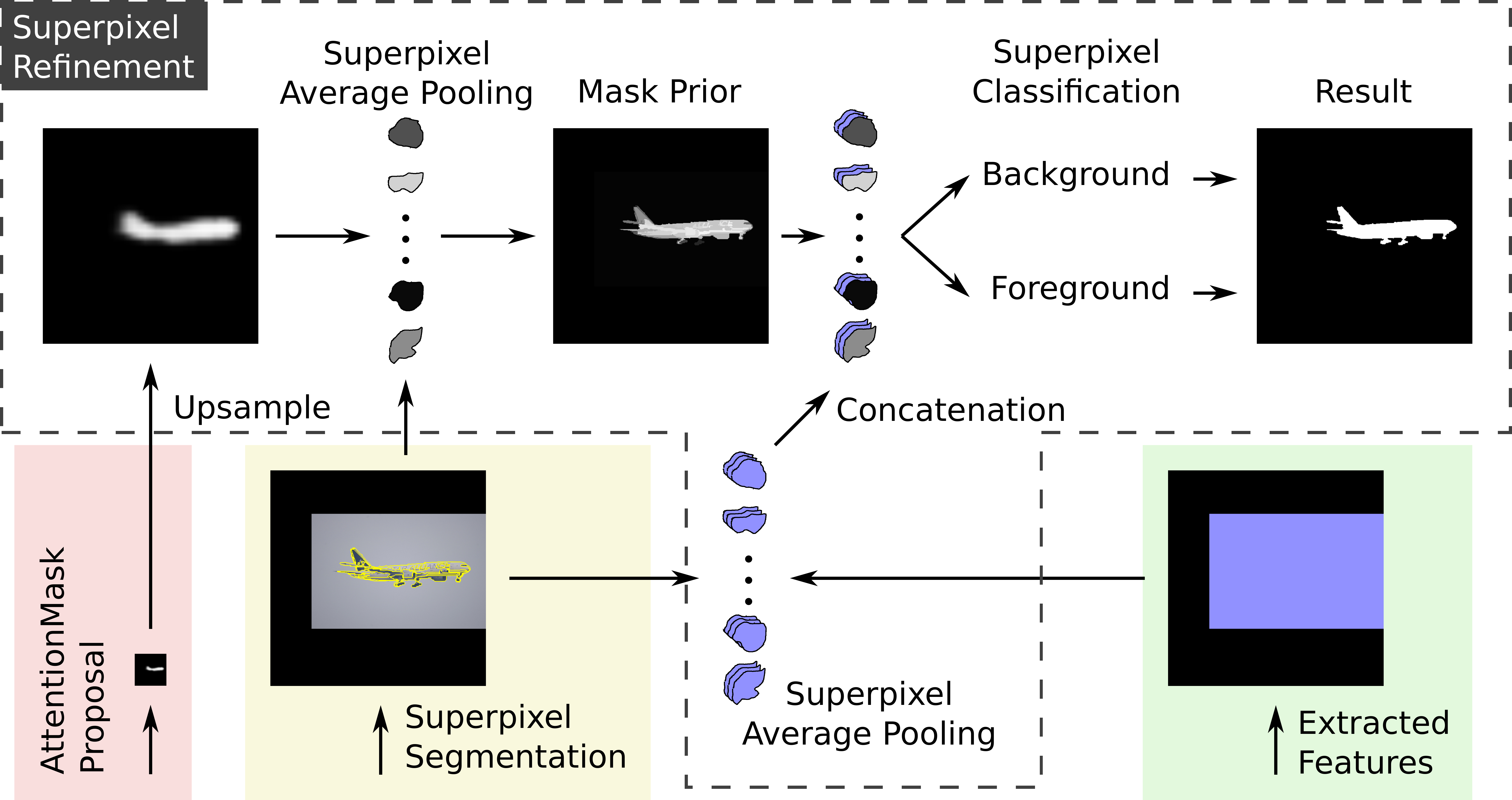}
\caption{Detailed view of our proposed superpixel refinement module showing the workflow for an individual proposal. The coarse AttentionMask proposal (red area) is upsampled and superpixel pooling is applied using the respective window from the segmentation stack (yellow area). This leads to a precise superpixelized version of the proposal called mask prior. Similarly, superpixel pooling is applied to the respective window of the feature stack (green area), yielding a feature vector per superpixel. Finally, mask prior and feature vector per superpixel are concatenated and classified as background or object part, leading to a precise proposal.}
\label{fig:ablaufSpxRefinement}
\end{figure*}

We apply the refinement, visualized based on an example in Fig.~\ref{fig:ablaufSpxRefinement}, to each proposal. The refinement starts by combining the upsampled AttentionMask proposal and the respective superpixel segmentation window. For each superpixel, the average probability, called mask prior, is computed using our superpixel pooling. This process is visualized in the upper left part of Fig~\ref{fig:ablaufSpxRefinement}. Utilizing the extracted features and the segmentations, a per superpixel feature vector is extracted applying our superpixel pooling again as visualized in the bottom part of Fig.~\ref{fig:ablaufSpxRefinement} (see Sec.~\ref{sec:meth_refinement_spxPooling}). Subsequently, mask prior and respective superpixel features are concatenated for each superpixel overlaying with the AttentionMask proposal window (upper central part of Fig.~\ref{fig:ablaufSpxRefinement}). Finally, these superpixels are classified with our new superpixel classifier (see Sec.~\ref{sec:meth_refinement_spxCls}) as background or object superpixels. The object superpixels constitutes the refined object proposal. This process is visualized in the top right of Fig.~\ref{fig:ablaufSpxRefinement}.  The final proposals combine the advantages of detailed superpixels, deep features and semantically rich but coarse AttentionMask object proposals. See Sec.~\ref{sec:meth_refinement_combination} for details of the refinement's integration with AttentionMask and DeepFH.

\subsection{Superpixel Pooling}
\label{sec:meth_refinement_spxPooling}
Following~\cite{he2017std2p,kwak2017weakly,park2017superpixel}, we propose a superpixel pooling module, to combine CNNs and superpixels. Directly integrating superpixels and CNNs is non-trivial, as outlined in the introduction. Superpixel pooling is similar to standard pooling, however, the final value is pooled from an arbitrarily shaped neighborhood (superpixel). Thus, the result has no lattice structure anymore. It is one feature or feature vector per superpixel, which has to be processed individually. Backpropagation is possible, similar to standard pooling. 

We utilize our superpixel pooling module twice in our refinement, as visualized in Fig.~\ref{fig:ablaufSpxRefinement}. First, we use the superpixel pooling to create the mask prior, a superpixelized version of the coarse AttentionMask proposal, shown in the upper left of Fig.\ref{fig:ablaufSpxRefinement}. The AttentionMask proposals are $40 \times 40$ windows representing parts of the image with different sizes in the original image. The proposals contain per-pixel probabilities for being part of an object. These windows are traced back to a level and spatial position of AttentionMask's feature pyramid by the attention system in AttentionMask. Given that level and the window's spatial position, we extract a crop from a suitable high-resolution superpixel segmentation, covering the same area in the input image as shown by the colored windows in Fig.~\ref{fig:sysFigSpxRefinement}. 

The superpixel segmentation's resolution depends on the level in AttentionMask's feature pyramid the proposal originates from. For proposals covering small objects, fine segmentations are used and vice versa. The segmentation crop's size is not fixed, but represents the input image area in high resolution. Thus, the segmentation can cover all details of the object. Given this segmentation crop and one proposal, we use superpixel pooling to create an average value of the proposal per superpixel. The value represents the probability of the superpixel belonging to the object. This is called mask prior and is applied individually per proposal.

Furthermore, we utilize our new superpixel pooling to compose a feature vector from the extracted features (green part in Fig.~\ref{fig:sysFigSpxRefinement}) for each superpixel of each level. Different from the mask prior computation described above, this computation is shared between all proposals from the same feature pyramid level, leading to an efficient execution. Per proposal, the mask prior and the feature vector for each superpixel overlaying with the proposal window are concatenated (see central part in Fig.~\ref{fig:ablaufSpxRefinement}). This constitutes a batch of superpixel representations per proposal for classification.

\subsection{Superpixel Classification}
\label{sec:meth_refinement_spxCls}
\begin{figure*}[t]
\centering
\includegraphics[width=\textwidth]{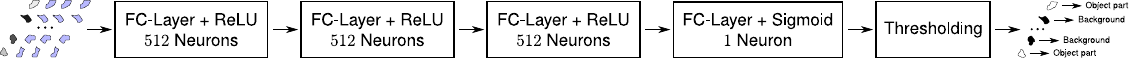}
\caption{Architecture of our superpixel classifier with fully-connected layers denoted as FC-layer. The input to the classifier is a batch of superpixels representing a proposal. For each superpixel, it consists of one mask prior value (gray) from the AttentionMask result and 512 learned features (bluish). The classifier determines if a superpixel is part of an object (white) or background (black).}
\label{fig:spxClassifier}
\end{figure*}

After applying superpixel pooling twice, each coarse AttentionMask proposal is a batch of superpixel feature vectors. These vectors represent precise superpixels with a resolution suitable for the proposal size. To determine the final refined proposal, each of the superpixel feature vectors is classified as background or object part (upper right in Fig.~\ref{fig:ablaufSpxRefinement}). For this classification, we propose a lightweight superpixel classifier visualized in Fig.~\ref{fig:spxClassifier} with only four fully-connected layers. The light-weight architecture is beneficial in terms of results compared to other structures as Tab.~\ref{tab:classifier} indicates. Altogether, the classifier takes a batch of superpixel features and returns the label object or background per superpixel. Superpixels that do not overlap with the proposal window are set as background to reduce computation. 

\begin{table}[t]
\renewcommand{\arraystretch}{1.2}
\centering
\caption{Architectures for the proposed superpixel classifier. The first column indicates the amount of neurons per fully-connected layer. AR@$n$ denotes the average recall for the first $n$ proposals.
}
\label{tab:classifier}
{
\begin{tabular}{@{}lcccccc@{}}
\hline
Architecture &  AR@10$\uparrow$ & AR@100$\uparrow$  & AR@1000$\uparrow$ \\ \hline
$\mathbf{512-512-512}$ & \textbf{0.104} &  \textbf{0.223} & \textbf{0.330} \\
$1024-1024-1024$ & 0.104 &  0.221 & 0.325  \\
$256-256-256$ & 0.102 &  0.220 & 0.327  \\
$512-512-512-512$ & 0.099 &  0.223 & 0.328  \\
$512-512$ & 0.101 &  0.224 & 0.328  \\
\hline
\end{tabular}
}
\end{table}

\subsection{Combination with AttentionMask and DeepFH}
\label{sec:meth_refinement_combination}
Combining the proposed superpixel-based refinement with AttentionMask into one end-to-end system leads to a change in the backbone. The original AttentionMask backbone, a 4-stage ResNet-50 is replaced by a 4-stage ResNet-34, to fit the memory restrictions on GPUs. To extract features from the backbone for per-superpixel features, we evaluated different backbone stages  (see  Tab.~\ref{tab:features}) and chose the \emph{res2} stage. The feature extractor itself is a $1 \times 1$ convolution. Altogether, AttentionMask, the feature extractor and the superpixel classifier are trained end-to-end in one system.

\begin{table}[t]
\renewcommand{\arraystretch}{1.2}
\centering
\caption{Results of the proposed refinement approach using features from different ResNet stages for feature extraction. AR@$n$ denotes the average recall for the first $n$ proposals.}
\label{tab:features}
{
\begin{tabular}{@{}lcccccc@{}}
\hline
ResNet-block &  AR@10$\uparrow$ & AR@100$\uparrow$  & AR@1000$\uparrow$ \\ \hline
\textit{res1} & 0.101 &  0.223 & 0.326 \\ 
\textbf{\textit{res2}} & \textbf{0.104} & \textbf{0.223} & \textbf{0.330}  \\ 
\textit{res3} & 0.103 &  0.224 & 0.323 \\ 
\textit{res4} & 0.098 &  0.221 & 0.324 \\ 
\hline
\end{tabular}
}
\end{table}

Outside this network, we generate DeepFH segmentations (see Sec.~\ref{sec:meth_deepfh}). As described in Sec.~\ref{sec:meth_deepfh_training}, we train DeepFH on the COCO dataset with LVIS annotations. Subsequently, DeepFH is used to generate one segmentation for each level of AttentionMask's feature pyramid. We optimize the parameter $t$ in FH as well as the parameter $\alpha$ from Eq.~\ref{eq:newFH} to generate on average 500 superpixels on the level of the largest objects and 8000 superpixels on the level of the smallest objects, with regular intermediate steps for the intermediate levels. Directly setting a desired number of superpixels is impossible for DeepFH. We present a comparison to other segmentation algorithms and numbers of superpixels in Sec.~\ref{sec:eval_lvis_abl_seg}.

Finally, after generating the refined proposals based on the DeepFH superpixels, we apply a three-step post-processing to each proposal. First, we use bilinear filtering on superpixel level, using the superpixels' mean colors. Second, to remove segmentation artifacts, we apply morphological opening and closing to the proposals. This removes or fills tiny antenna-like artifacts in the proposal masks. Finally, we apply non-maximum suppression (NMS) with a high IoU of $0.95$ to remove almost identical proposals. Since all proposals are represented by few superpixels, identical proposals are more common than with pixel-based proposals. The post-processing is evaluated in Sec.~\ref{sec:eval_lvis_abl_pp}.

\subsection{Training}
\label{sec:meth_refinement_training}
We train our superpixel-based refinement with AttentionMask end-to-end using SGD with a learning rate of $0.0001$ for 13 epochs. The ResNet-34 backbone is initialized with ImageNet weights, while the rest of the network is learned from scratch. Apart from AttentionMask, the feature extractor and the superpixel classifier are trained. The superpixel classifier is trained with a binary crossentropy loss ($\mathcal{L}_{spx}$), while the feature extractor is automatically trained by the overall system. Thus, it does not require distinct supervision. Combined with the AttentionMask~\cite{WilmsFrintropACCV2018} losses $\mathcal{L}_{objn}, \mathcal{L}_{ah}, \mathcal{L}_{seg}$, and $\mathcal{L}_{att_n}$, the overall loss is defined as
\begin{equation}
\begin{split}
\mathcal{L} = & w_{objn}\mathcal{L}_{objn}+w_{ah}\mathcal{L}_{ah}+w_{seg}\mathcal{L}_{seg} + \\
&\quad w_{att} \sum_n \mathcal{L}_{att_n} + w_{spx} \mathcal{L}_{spx}
\end{split}
\end{equation}
with weights controlling the losses' influence. $w_{spx}=1$ throughout our experiments, while the other weights stay unchanged compared to~\cite{WilmsFrintropACCV2018}.

As training data we use the training images of the COCO dataset~\cite{lin2014microsoft} with the COCO annotation for a fair comparison. The ground truth contains polygon annotations for objects of 80 classes. However, generating ground truth for the superpixel classifier is non-trivial, since the annotation boundaries do not necessarily match the superpixel boundaries. Thus, for each annotated object, we generate an optimal set of superpixels yielding a maximum IoU between the set and the ground truth annotation. This process starts from all superpixels completely contained in the annotated object and greedily adds superpixels that increase the IoU. These superpixels are positive samples, while all other superpixels form a pool of negative samples. 

\section{Evaluation}
\label{sec:eval}
For evaluating our new superpixel-based refinement approach combined with our novel DeepFH segmentations on top of AttentionMask, we mostly follow standard object proposal evaluation pipelines. We assess the quality of the results based on average recall (AR), which combines recall as well as the proposals' segmentation quality and is frequently used in object proposal literature~\cite{Hosang2015PAMI,Pinheiro2015-deepmask,Pinheiro2016-sharpmask,Hu2017-fastmask,WilmsFrintropACCV2018}. AR is calculated for different numbers of proposals, e.g., AR@100 for the first 100 proposals, and for different sizes of objects. To only assess the segmentation quality of the proposals, we use standard segmentation measures boundary recall (BR) undersegmentation error (UE), and oversegmentation error (OE).

Finally, we introduce the \underline{A}chievable \underline{I}ntersection \underline{o}ver \underline{U}nion (AIoU) as a new measure to assess the quality of a segmentation for superpixel-based object proposal generation. Since the ground truth for our system was superpixelized for training (see Sec.~\ref{sec:meth_refinement_training}), we measure given a superpixel segmentation the average IoU between the optimal set of superpixels per annotated object and the annotated object itself. If the superpixel segmentation perfectly captures the annotated object, the AIoU is 1. Otherwise, the AIoU decreases to a value between 0 and 1. Thus, the AIoU is an upper bound for the quality of a superpixel-based proposal algorithm in terms of average IoU (AVGIoU) per ground truth object. Comparing AIoU and AVGIoU allows us to assess how much of the segmentation quality is utilized.

The dataset for our evaluation is the COCO~\cite{lin2014microsoft} dataset with more detailed LVIS~\cite{gupta2019lvis} annotations. This is different to other works~\cite{Pinheiro2015-deepmask,Pinheiro2016-sharpmask,Hu2017-fastmask,WilmsFrintropACCV2018}. However, it is important for our evaluation, since we focus on precise object proposals. Thus, we use the LVIS data for validation and testing, while all approaches were trained on COCO data. An evaluation of our baseline AttentionMask compared to other methods on COCO is presented in~\cite{WilmsFrintropACCV2018}. We compare our method to the CNN-based methods DeepMask~\cite{Pinheiro2015-deepmask}, SharpMask~\cite{Pinheiro2016-sharpmask}, FastMask~\cite{Hu2017-fastmask}, and AttentionMask~\cite{WilmsFrintropACCV2018} described in Sec.~\ref{sec:relWork_proposals} as well as the method MCG~\cite{pont2017multiscale}, which does not utilize CNNs and does not suffer from CNN-based downsampling.

\begin{table*}[t]
\renewcommand{\arraystretch}{1.2}
\centering
\caption{Results on the COCO dataset with LVIS annotations using average recall (AR). AR$^S$, AR$^M$ and AR$^L$ denote results on small, medium and large objects. Best and second-best results marked with \textbf{bold font} and \textit{italic font} respectively.}
\label{tab:resultsLVIS}
\begin{tabular}{@{}lccccccc@{}}
\hline
Method & Backbone &  AR@10$\uparrow$ & AR@100$\uparrow$  & AR@1000$\uparrow$ & AR$^S$@100$\uparrow$  & AR$^M$@100$\uparrow$  & AR$^L$@100$\uparrow$ \\ \hline 
MCG~\cite{pont2017multiscale} & - & 0.048  &  0.131 & 0.237  & 0.031  & 0.204  &  0.462 \\ \hline 
DeepMask~\cite{Pinheiro2015-deepmask} & ResNet-50 & 0.069  &  0.147 & 0.214  & 0.014  & 0.314  &  0.430\\ 
SharpMask~\cite{Pinheiro2016-sharpmask} & ResNet-50 & 0.073  &  0.154 & 0.229  & 0.014  & 0.327  &  0.460\\ 
FastMask~\cite{Hu2017-fastmask} & ResNet-50 & 0.069 &  0.161 & 0.256  & 0.055  & 0.296 & 0.386 \\ 
AttentionMask~\cite{WilmsFrintropACCV2018} & ResNet-50 & 0.073 &  0.189 & 0.284  & 0.081  & 0.312  & 0.446 \\ \hline 
AttentionMask~\cite{WilmsFrintropACCV2018} & ResNet-34  & 0.076 &  0.185 & 0.271  & 0.083  & 0.305  & 0.423 \\ 
Ours with FH~\cite{felzenszwalb2004efficient} & ResNet-34  & \textit{0.092} &  \textit{0.206} & \textit{0.290} & \textit{0.092} & \textit{0.340} & \textit{0.473 } \\ 
Ours with DeepFH & ResNet-34  & \textbf{0.094} &  \textbf{0.213} & \textbf{0.304} & \textbf{0.098} & \textbf{0.349} & \textbf{0.480} \\ 
\hline
\end{tabular}
\end{table*}

\subsection{Results on LVIS}
\label{sec:eval_lvis}
First, we present the results of our proposed system on the object proposal generation task using the COCO dataset with LVIS annotations. From the quantitative results in Tab.~\ref{tab:resultsLVIS} it is visible that our proposed superpixel-based refinement with FH or DeepFH outperforms all other methods. In terms of AR@100, we surpass the other methods by $0.024$ to $0.082$. Thus, the refinement improves the results despite using a smaller backbone. For instance, the results of AttentionMask using a ResNet-34 backbone are slightly worse compared to original AttentionMask. Our refinement improves the results significantly even surpassing original AttentionMask as well as FastMask, SharpMask, DeepMask and MCG.

Comparing the numbers for objects of different sizes, we can show an improvement across all object sizes. Interestingly, for large objects (AR$^L$@100) SharpMask and the not DL-based MCG outperform original AttentionMask. This could not be shown using the COCO annotations~\cite{WilmsFrintropACCV2018}. However, it is not surprising, since FastMask and AttentionMask heavily utilize subsampling. Yet, our proposed refinement improves AttentionMask's results for large objects, outperforming SharpMask and MCG. Finally, comparing the bottom rows, it is visible that our DeepFH segmentations improve the results compared to classic FH segmentations. We discuss this difference in more detail in our ablation studies.

\begin{figure*}[t]
\centering
\begin{tabular}{ccccc}
\includegraphics[width = .175\linewidth]{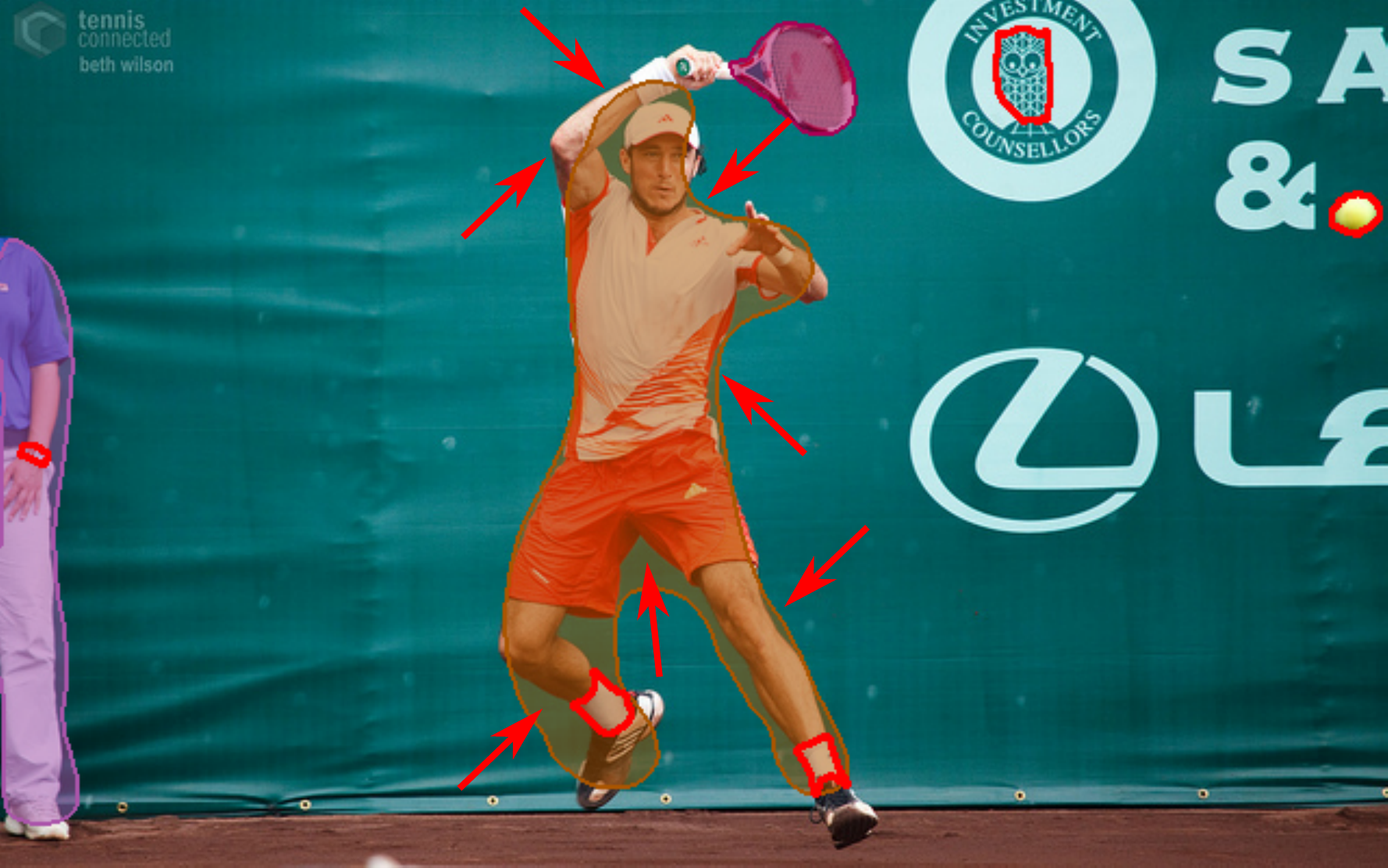} &
\includegraphics[width = .175\linewidth]{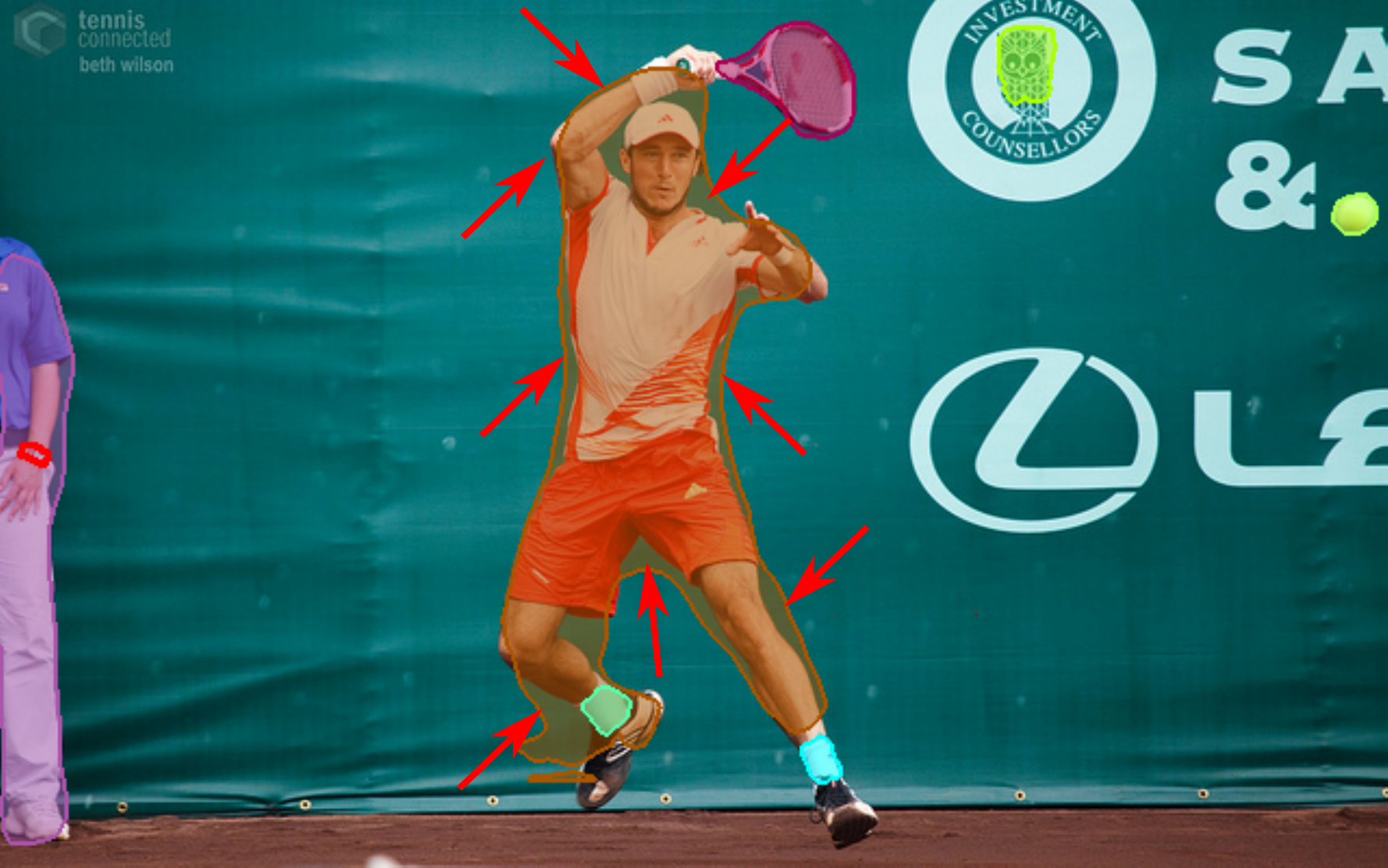} &
\includegraphics[width = .175\linewidth]{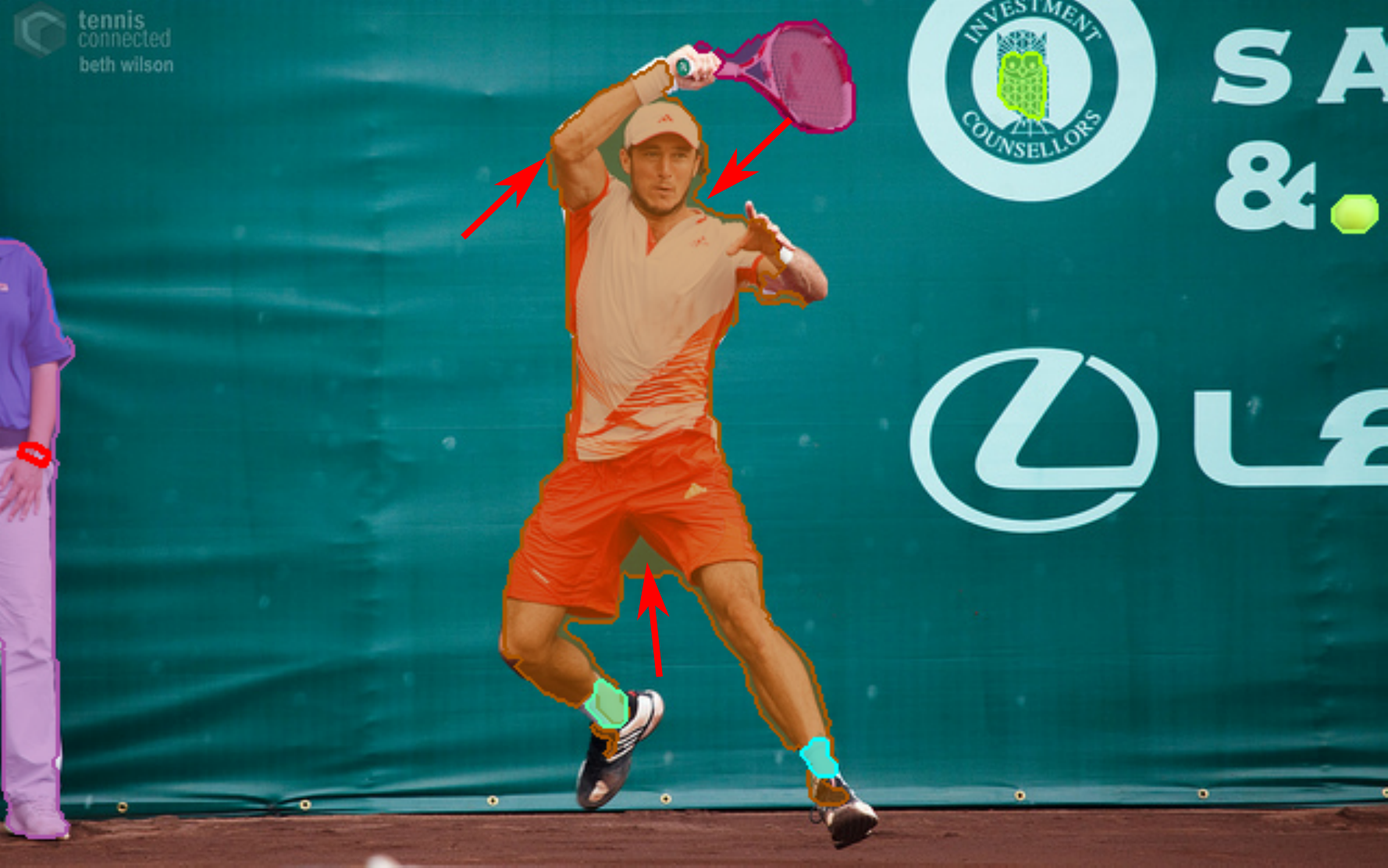} &
\includegraphics[width = .175\linewidth]{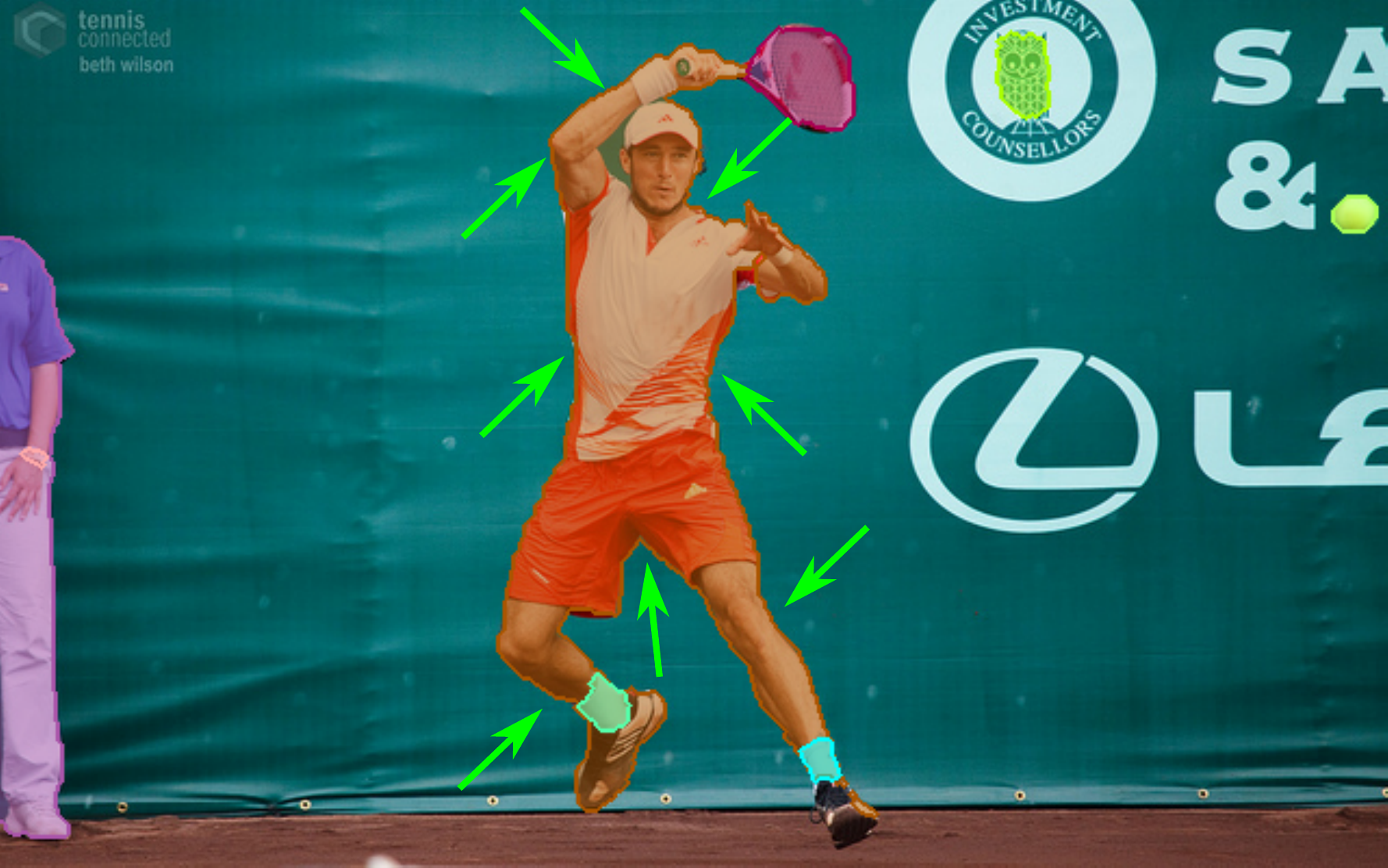} &
\includegraphics[width = .175\linewidth]{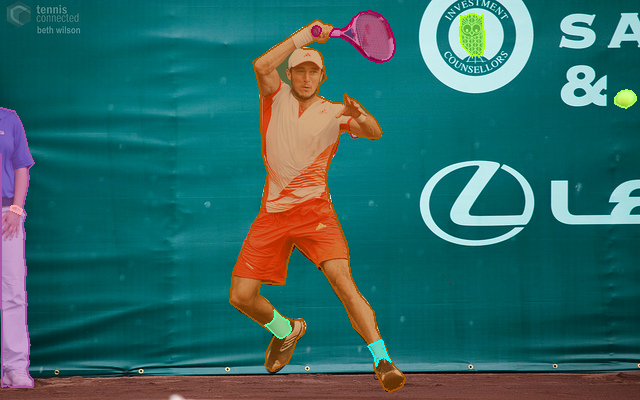}
\\ 
\includegraphics[width = .175\linewidth]{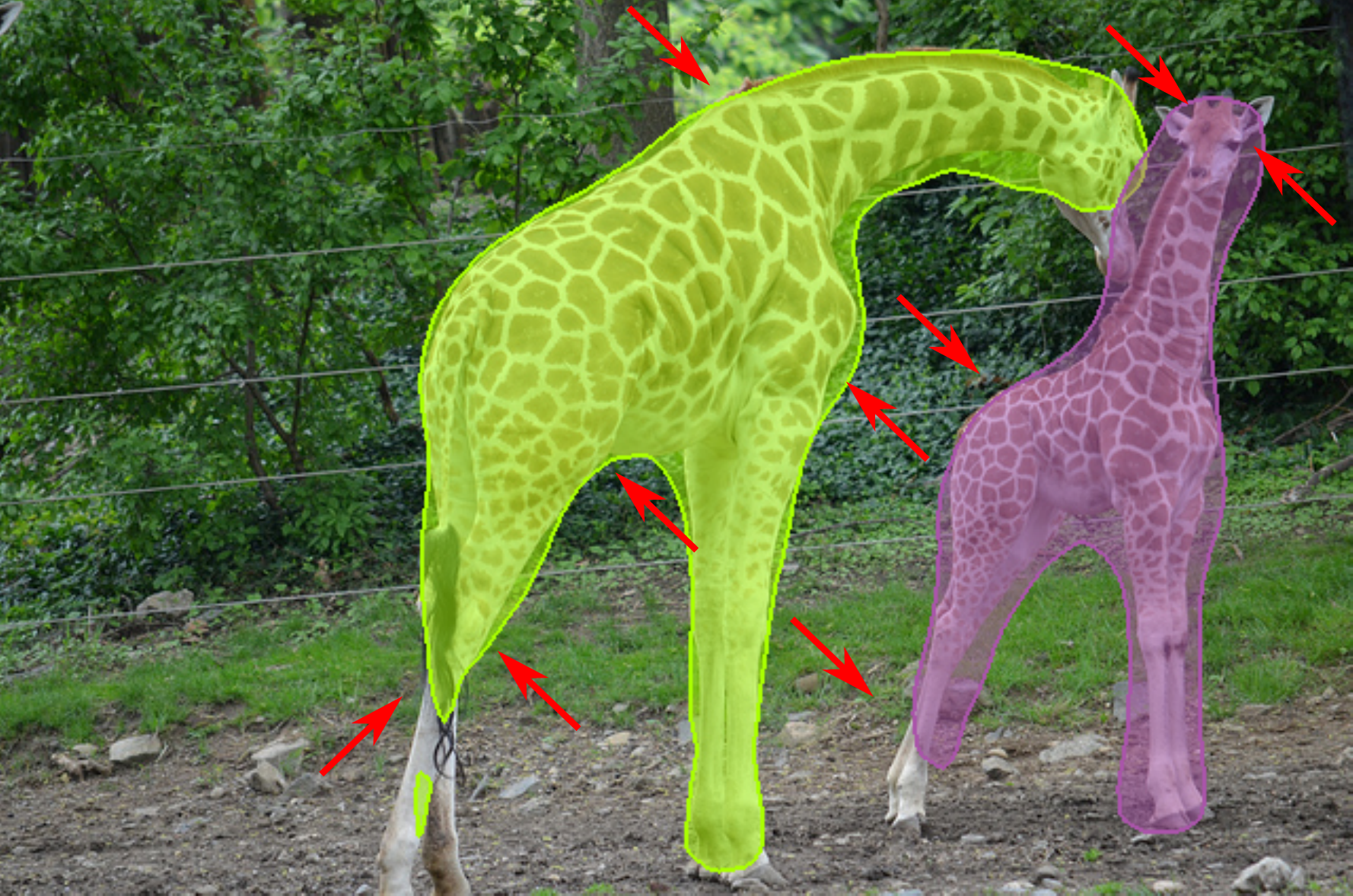} &
\includegraphics[width = .175\linewidth]{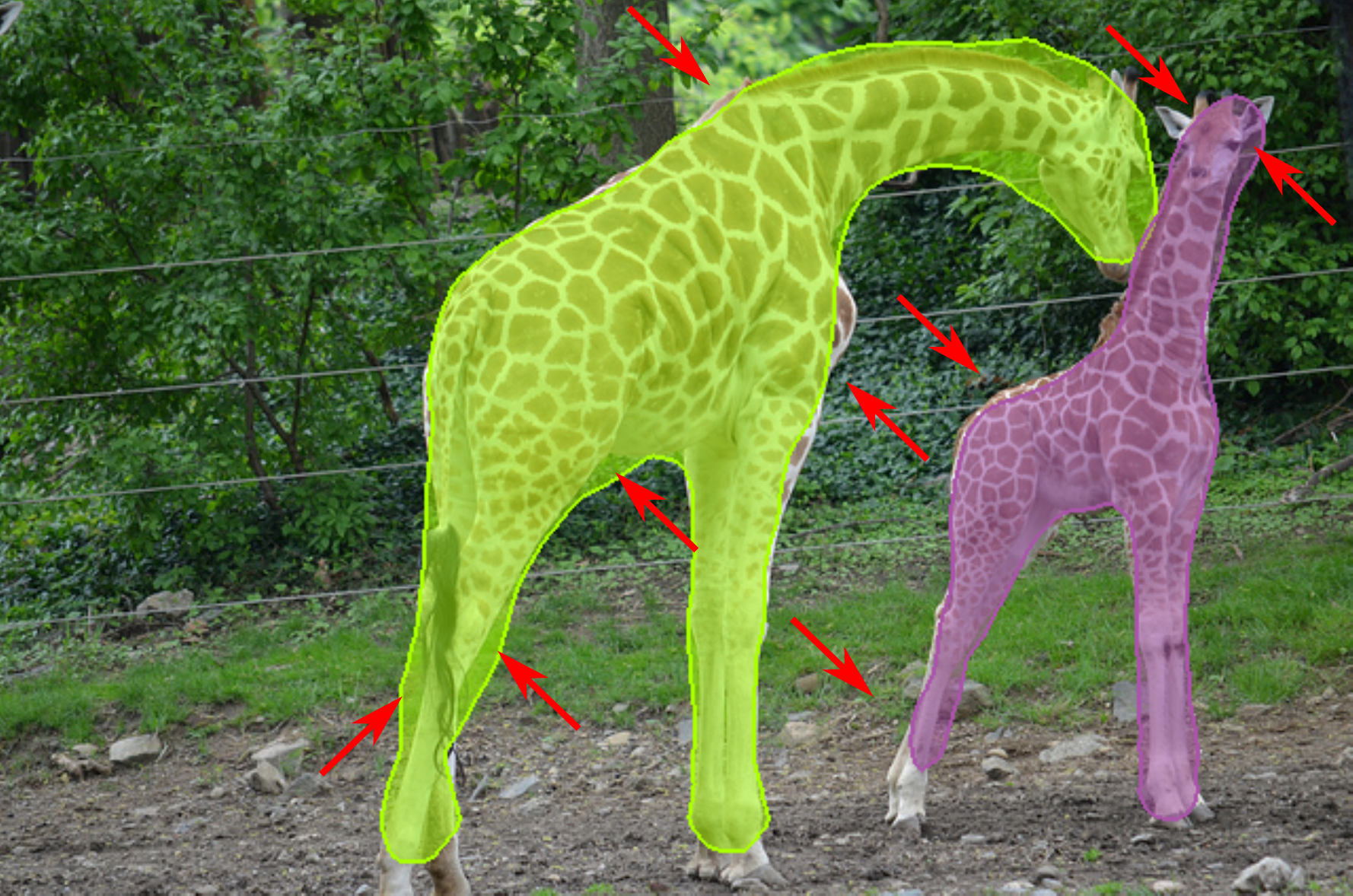} &
\includegraphics[width = .175\linewidth]{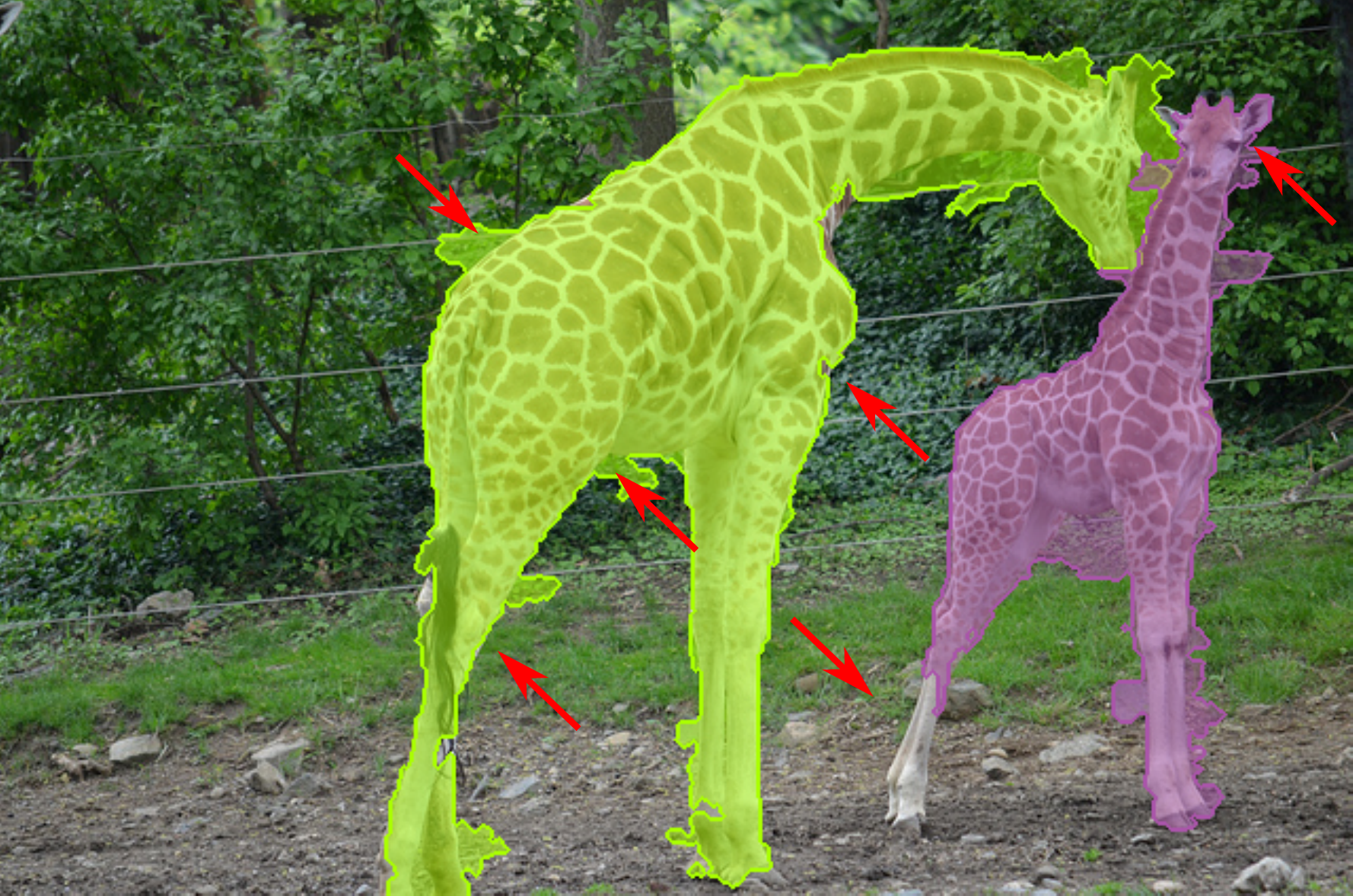} &
\includegraphics[width = .175\linewidth]{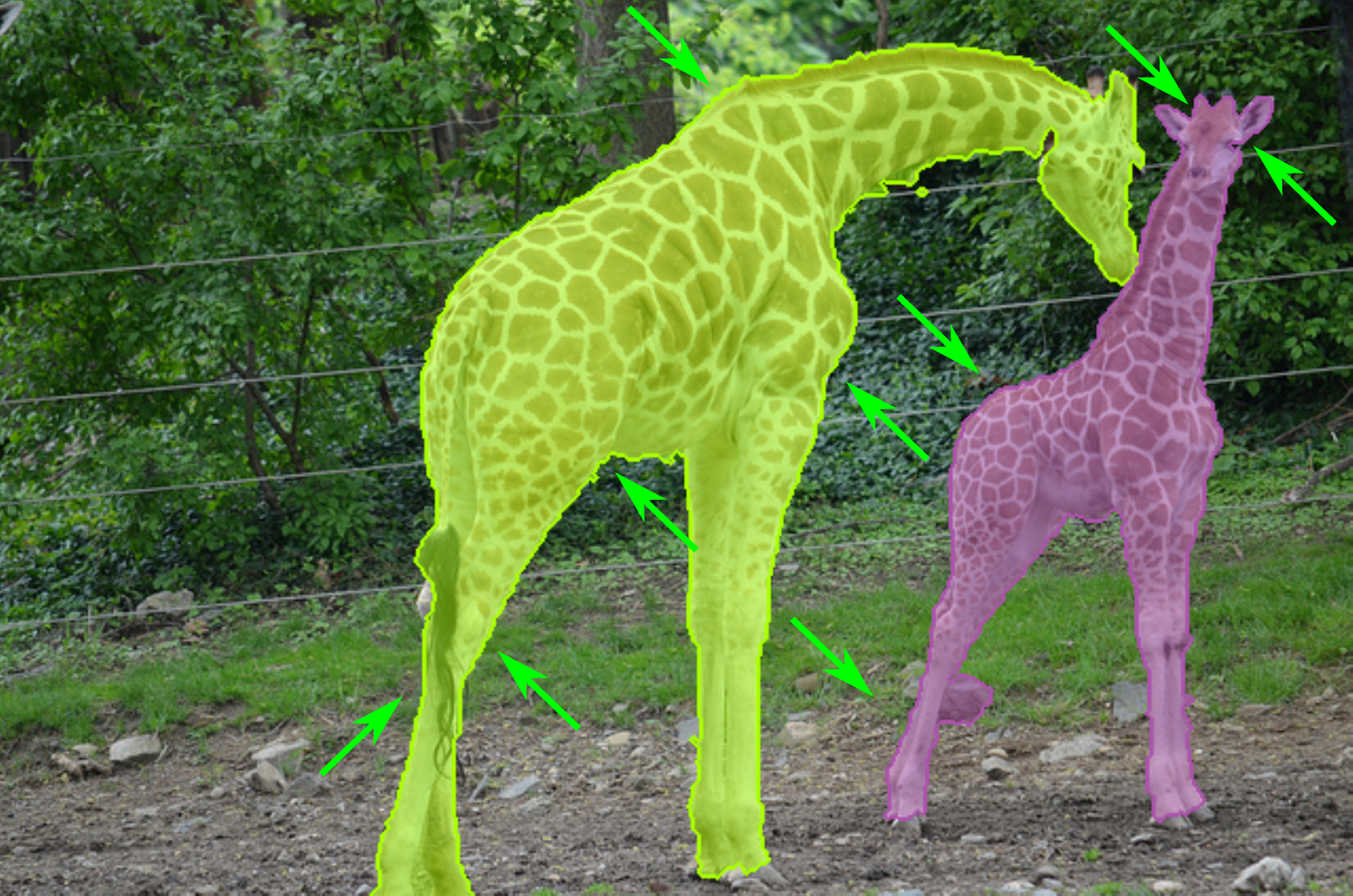} &
\includegraphics[width = .175\linewidth]{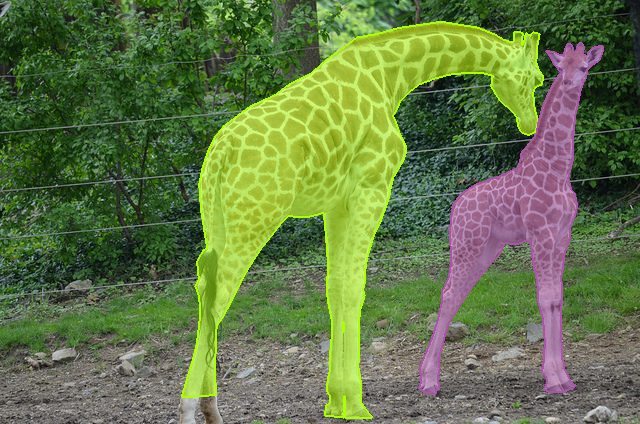}
\\
\includegraphics[width = .175\linewidth]{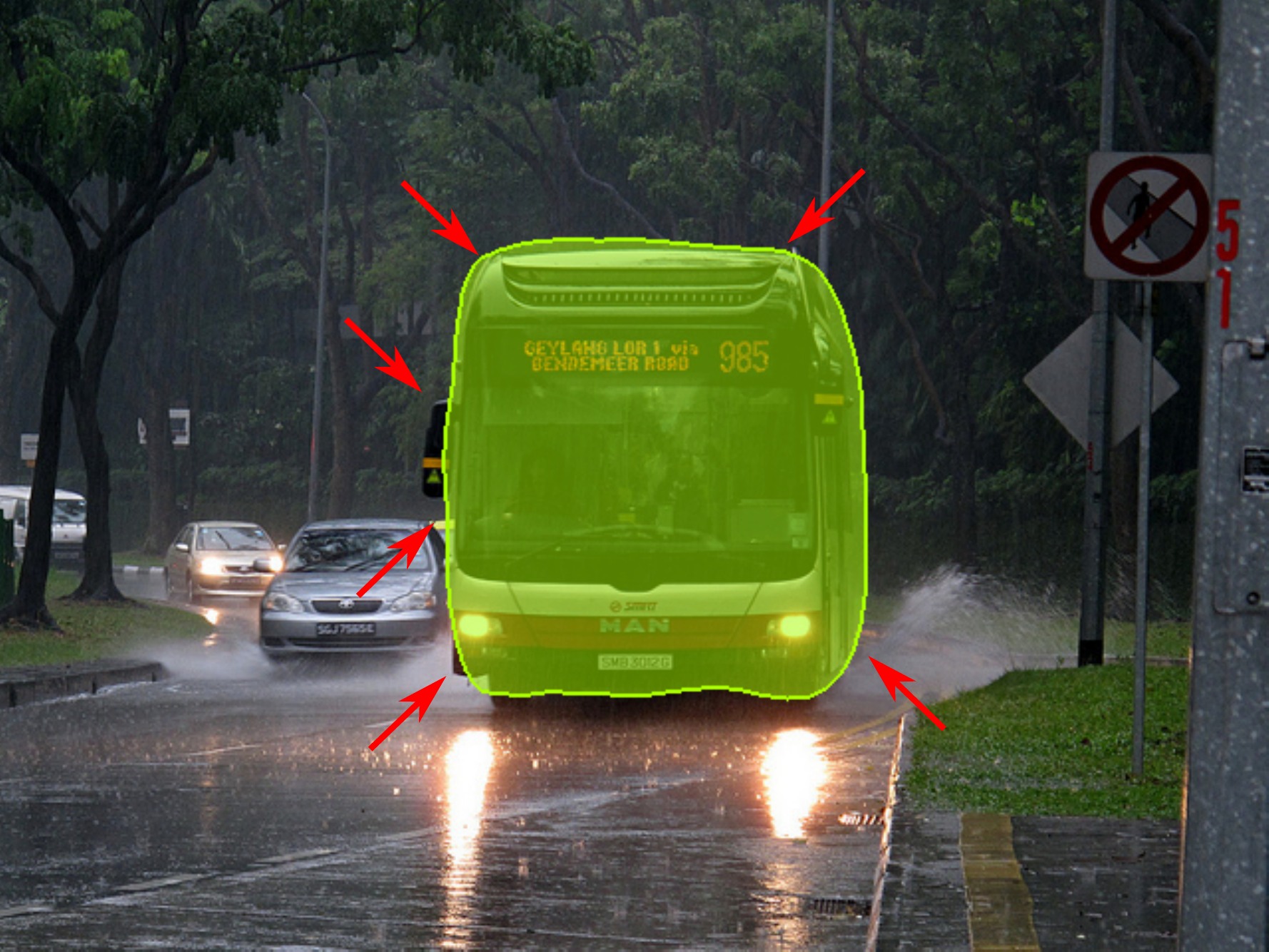} &
\includegraphics[width = .175\linewidth]{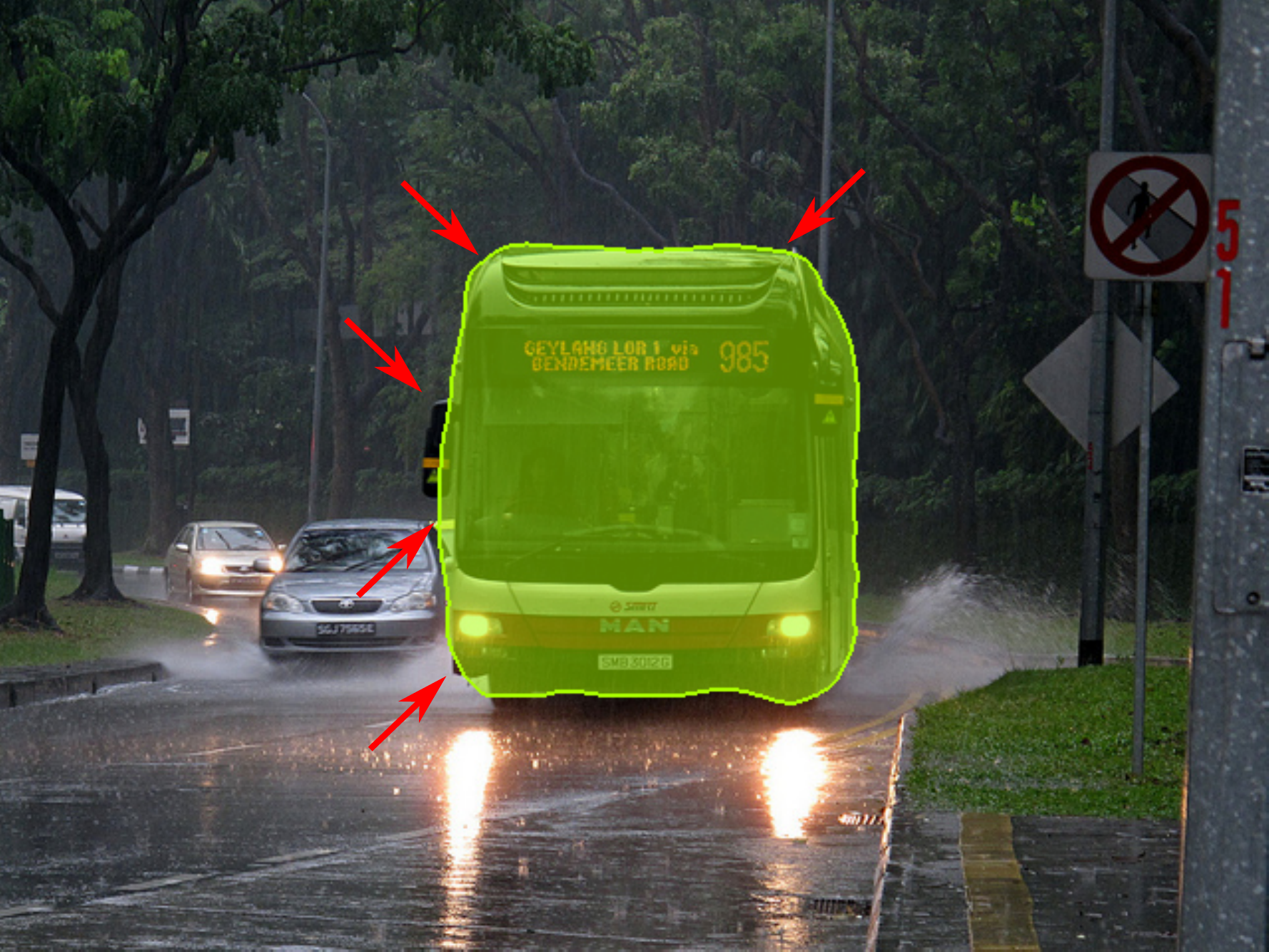} &
\includegraphics[width = .175\linewidth]{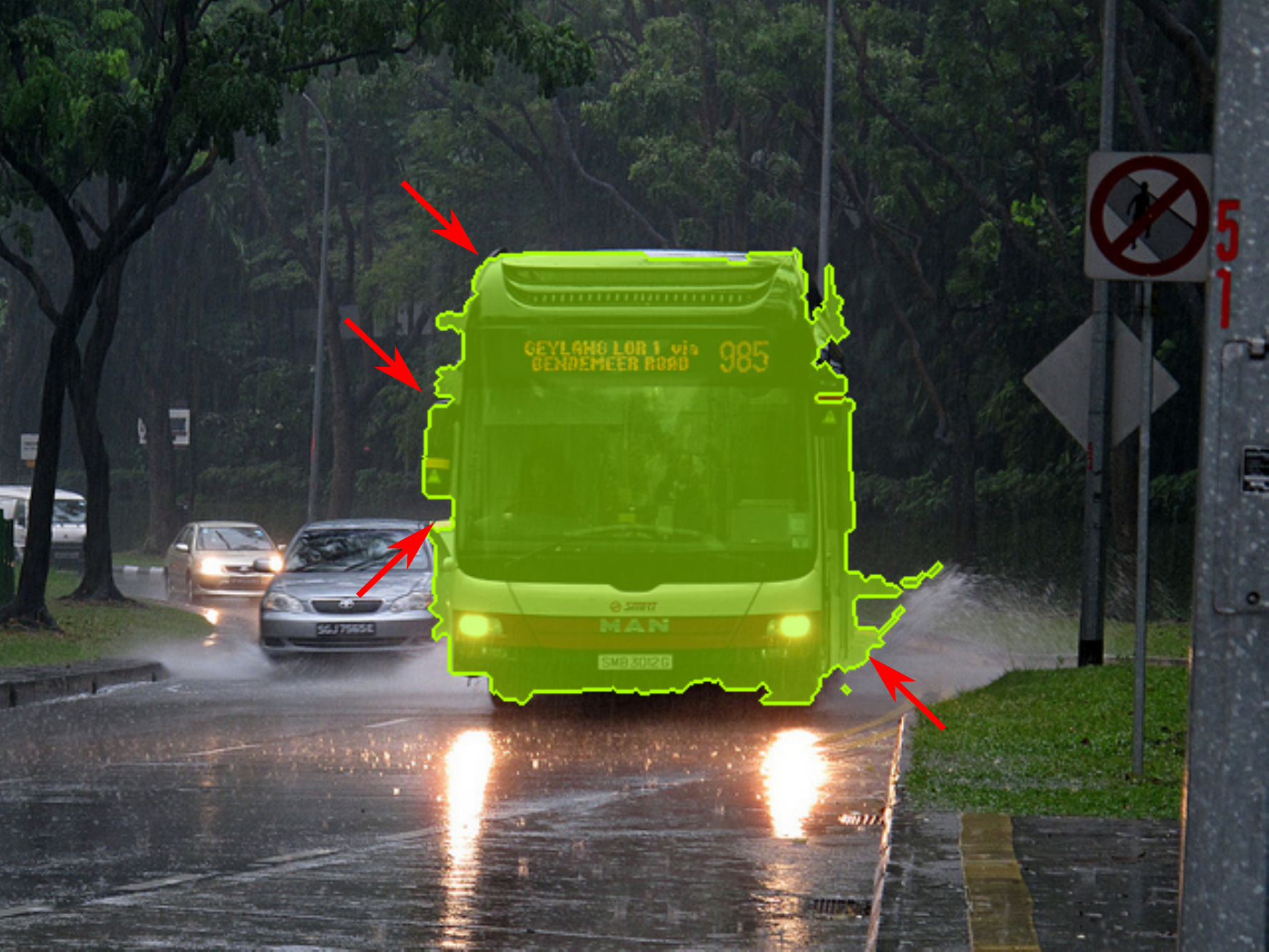} &
\includegraphics[width = .175\linewidth]{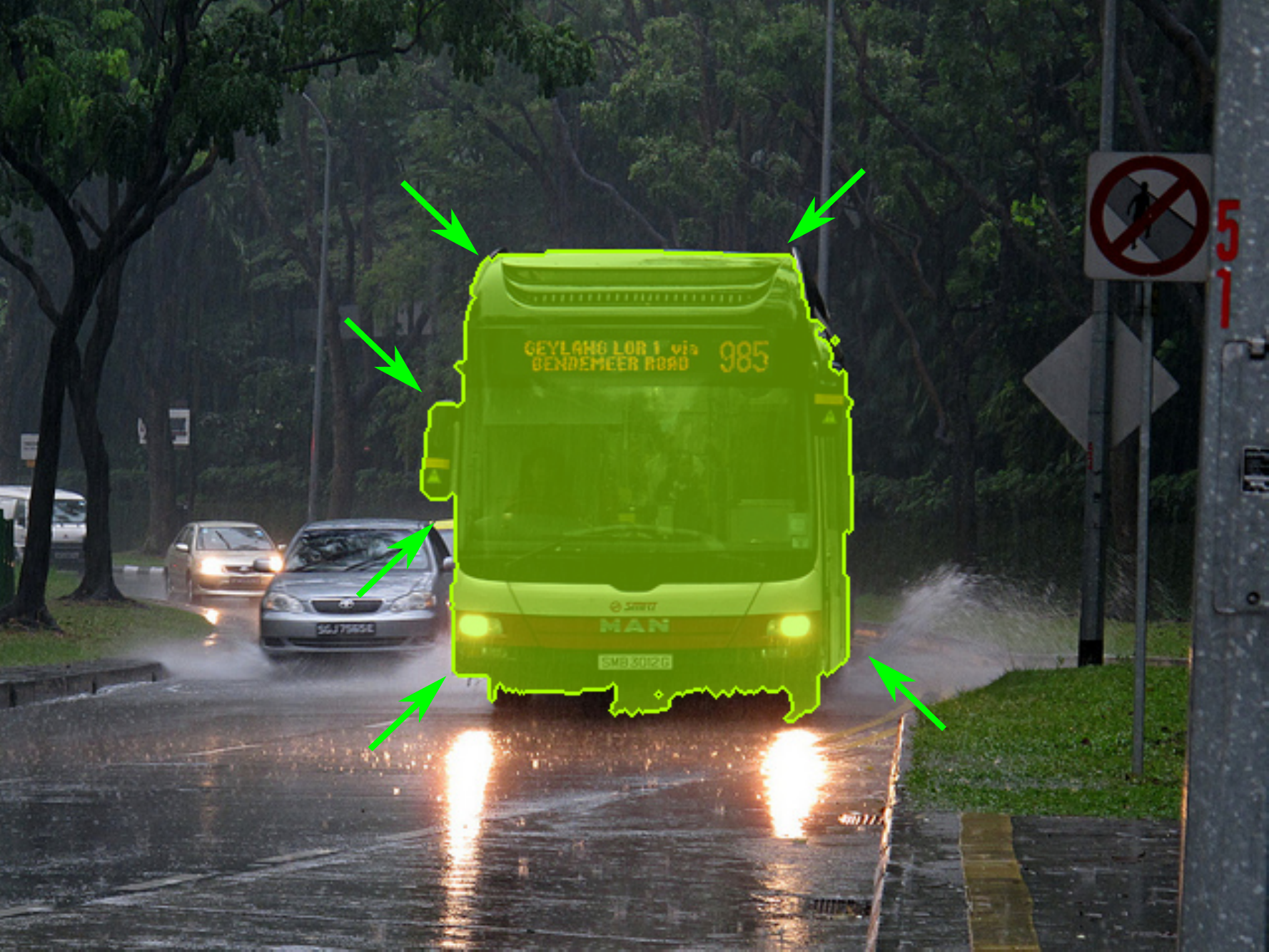} &
\includegraphics[width = .175\linewidth]{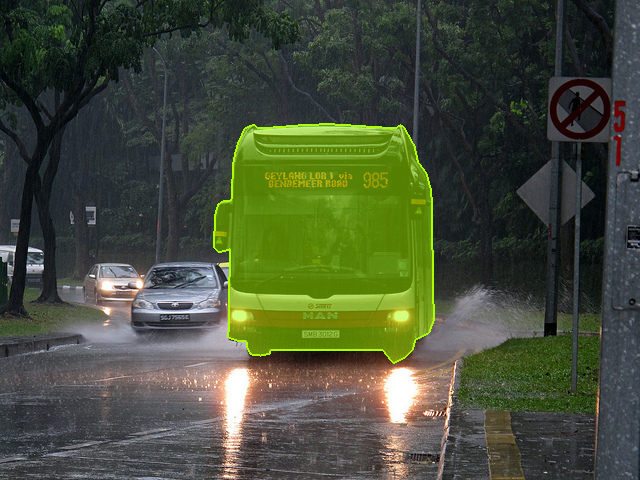} 
\\
\includegraphics[width = .175\linewidth]{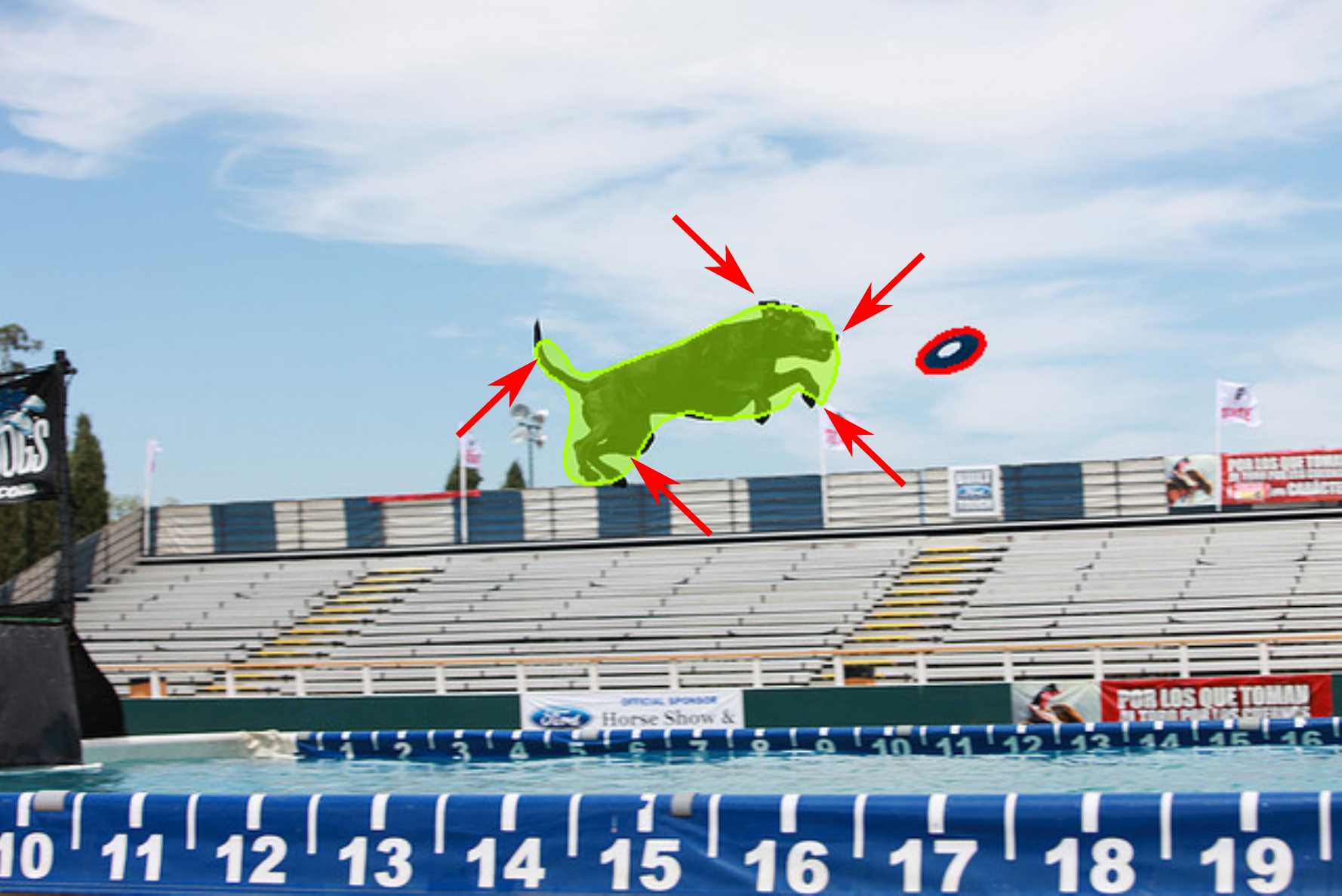} &
\includegraphics[width = .175\linewidth]{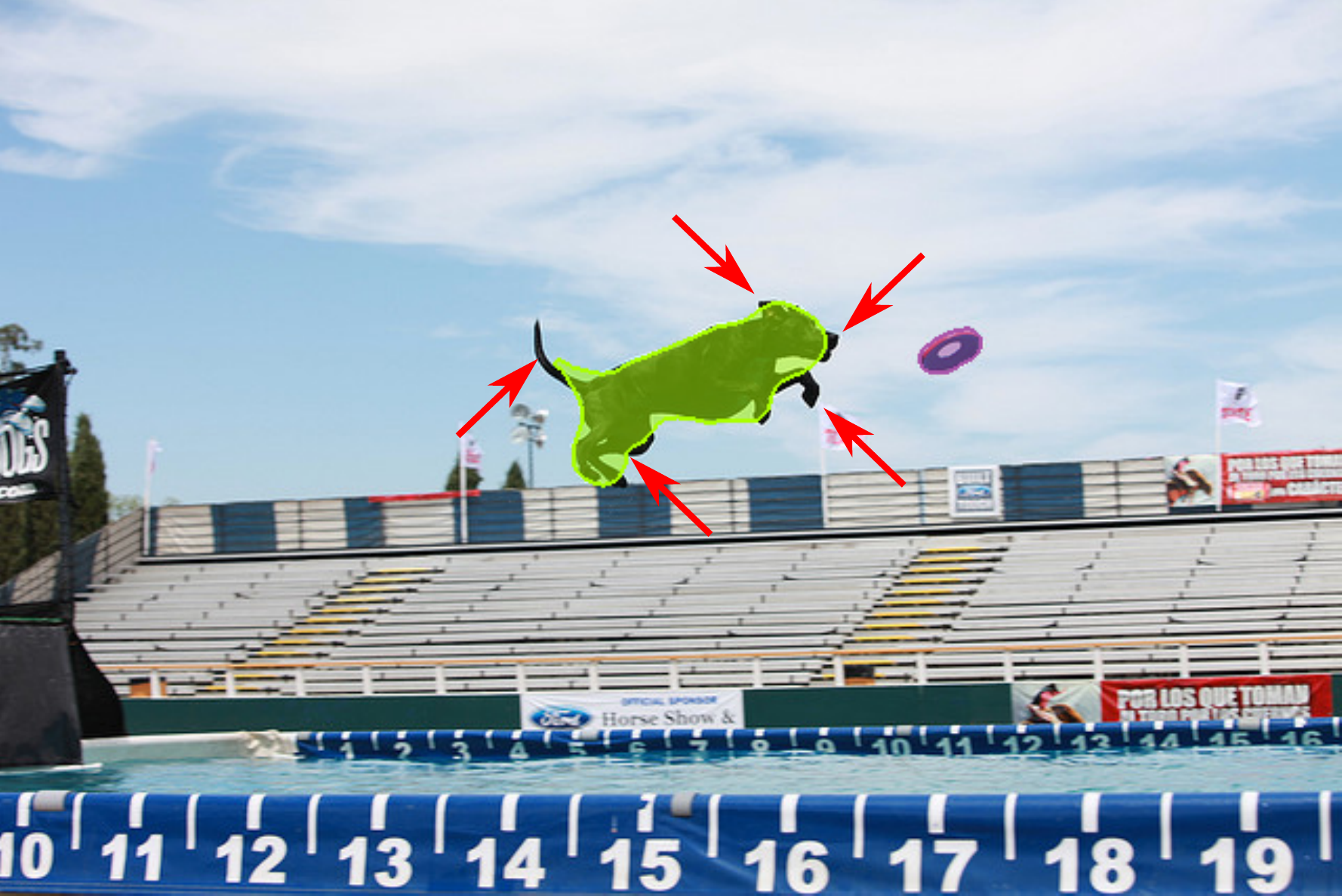} &
\includegraphics[width = .175\linewidth]{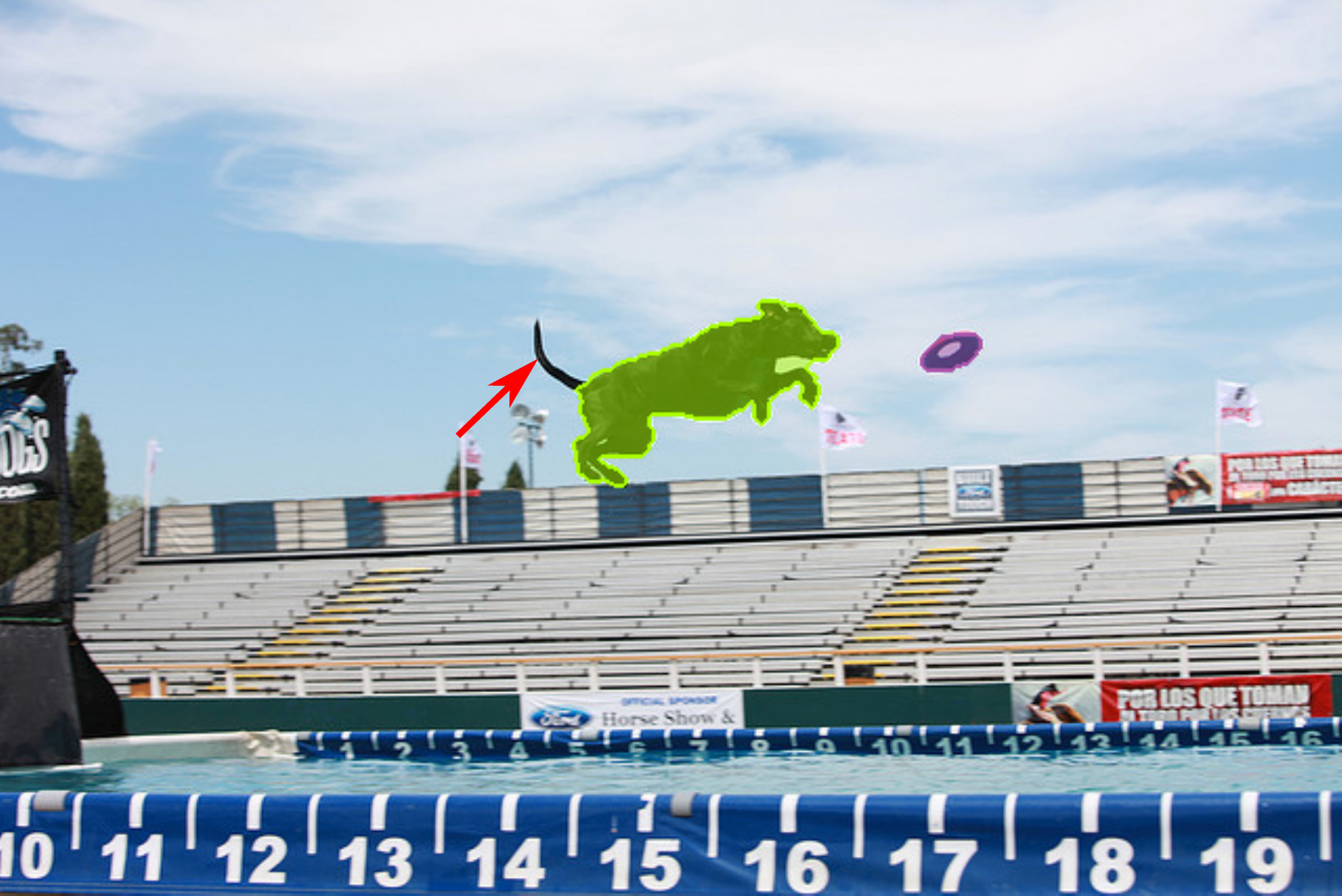} &
\includegraphics[width = .175\linewidth]{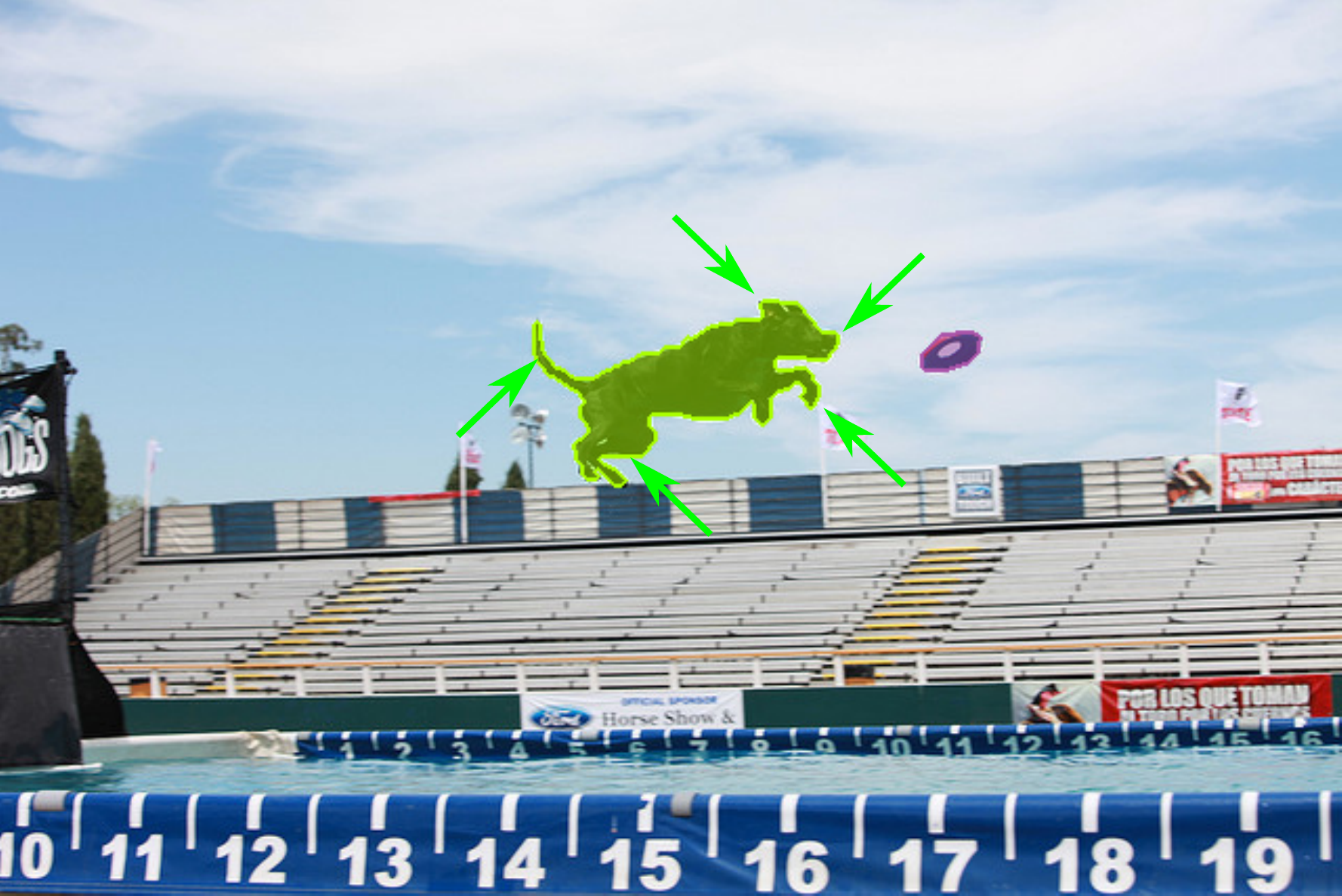} &
\includegraphics[width = .175\linewidth]{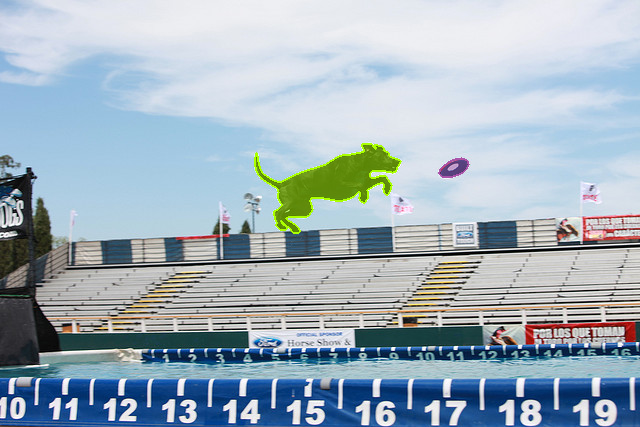} 
\\
\includegraphics[width = .175\linewidth]{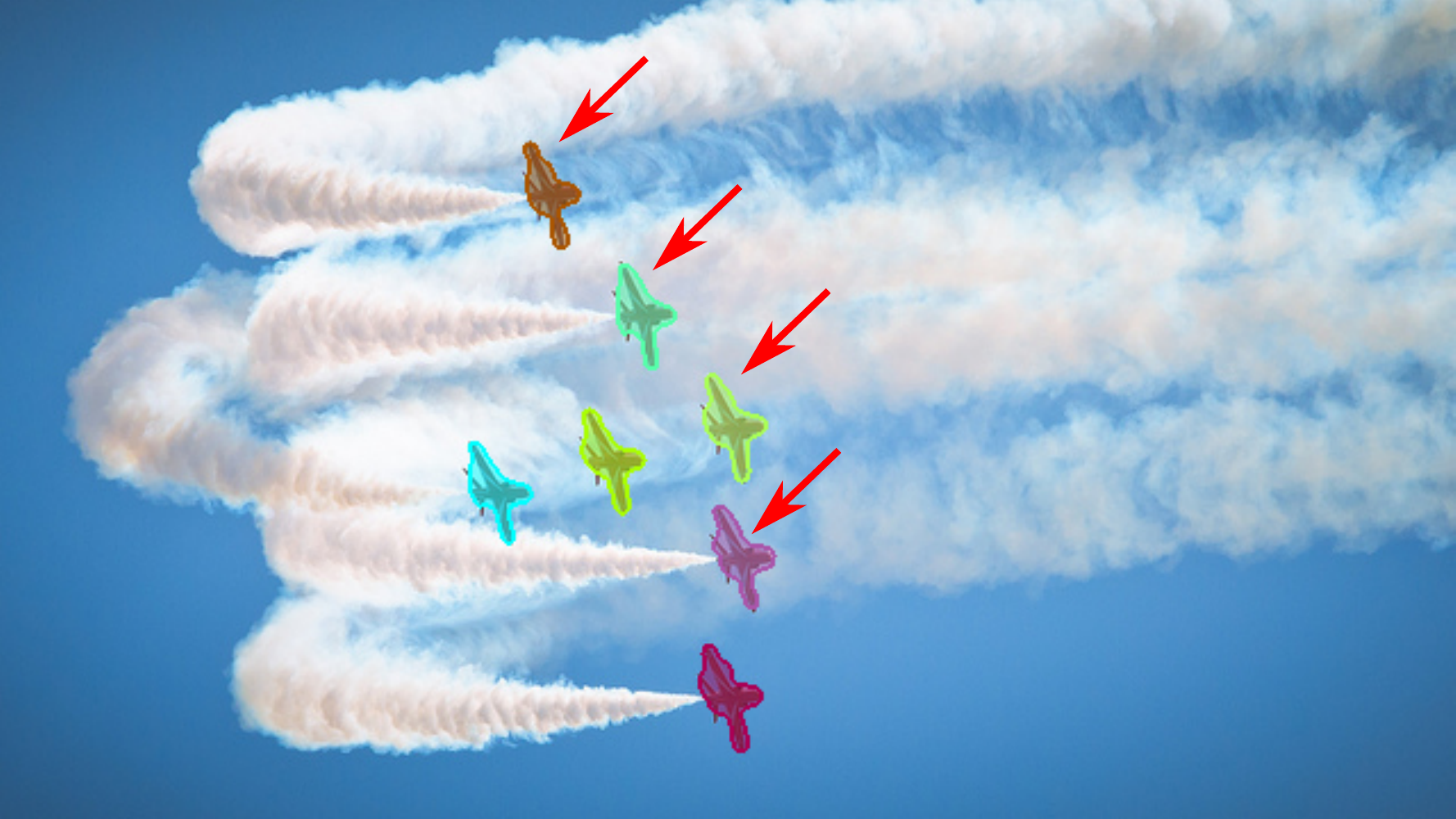} &
\includegraphics[width = .175\linewidth]{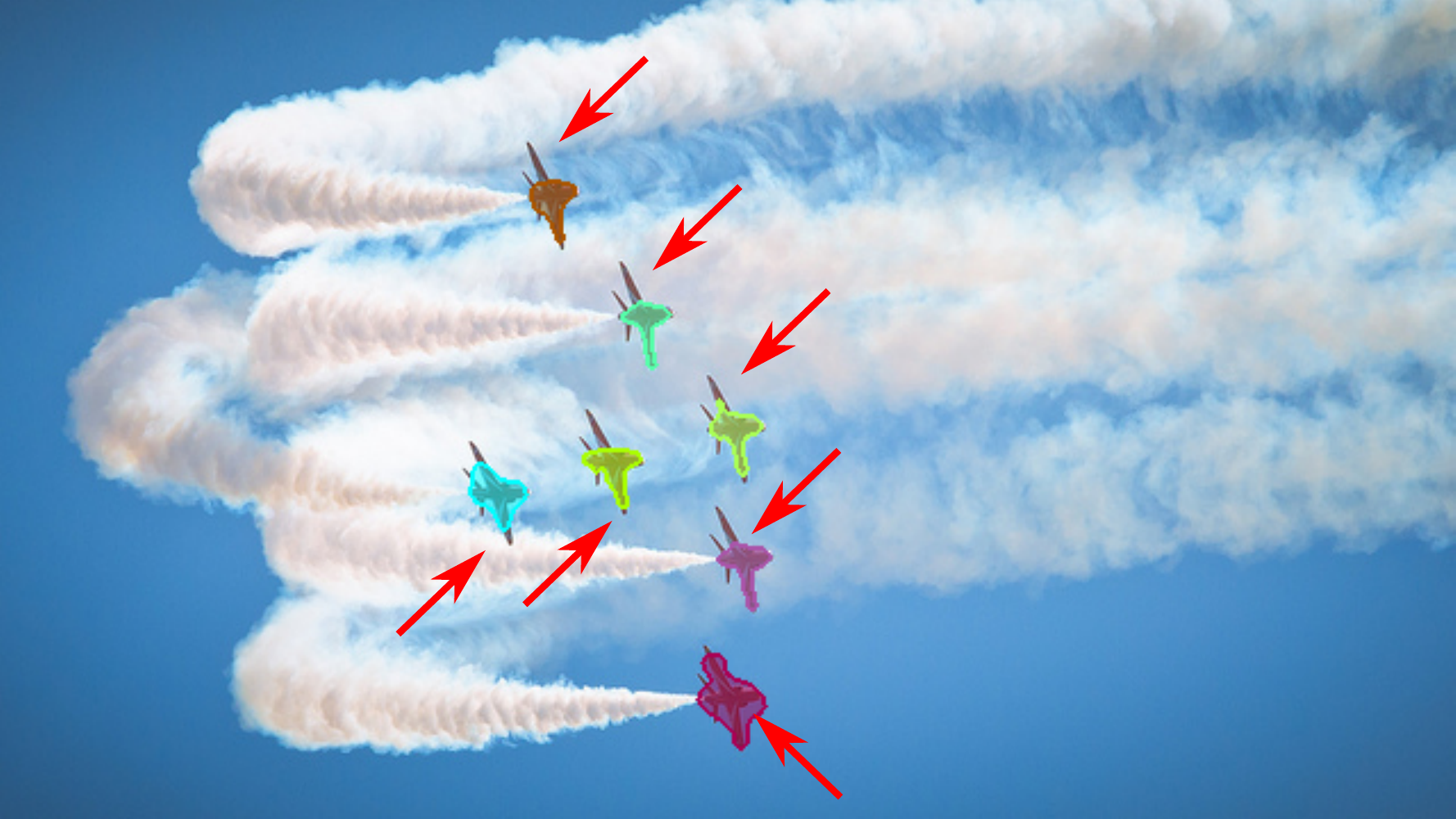} &
\includegraphics[width = .175\linewidth]{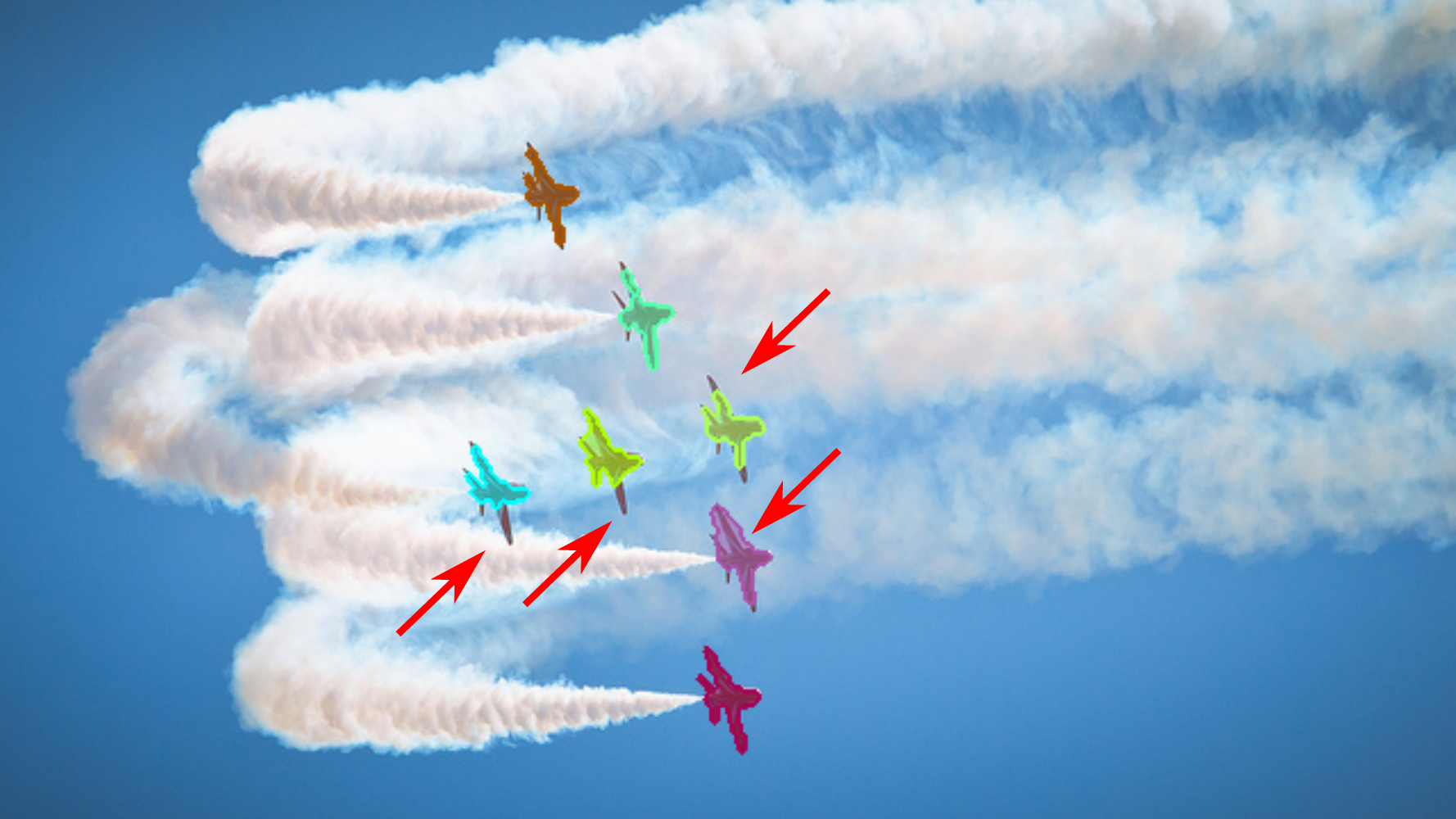} &
\includegraphics[width = .175\linewidth]{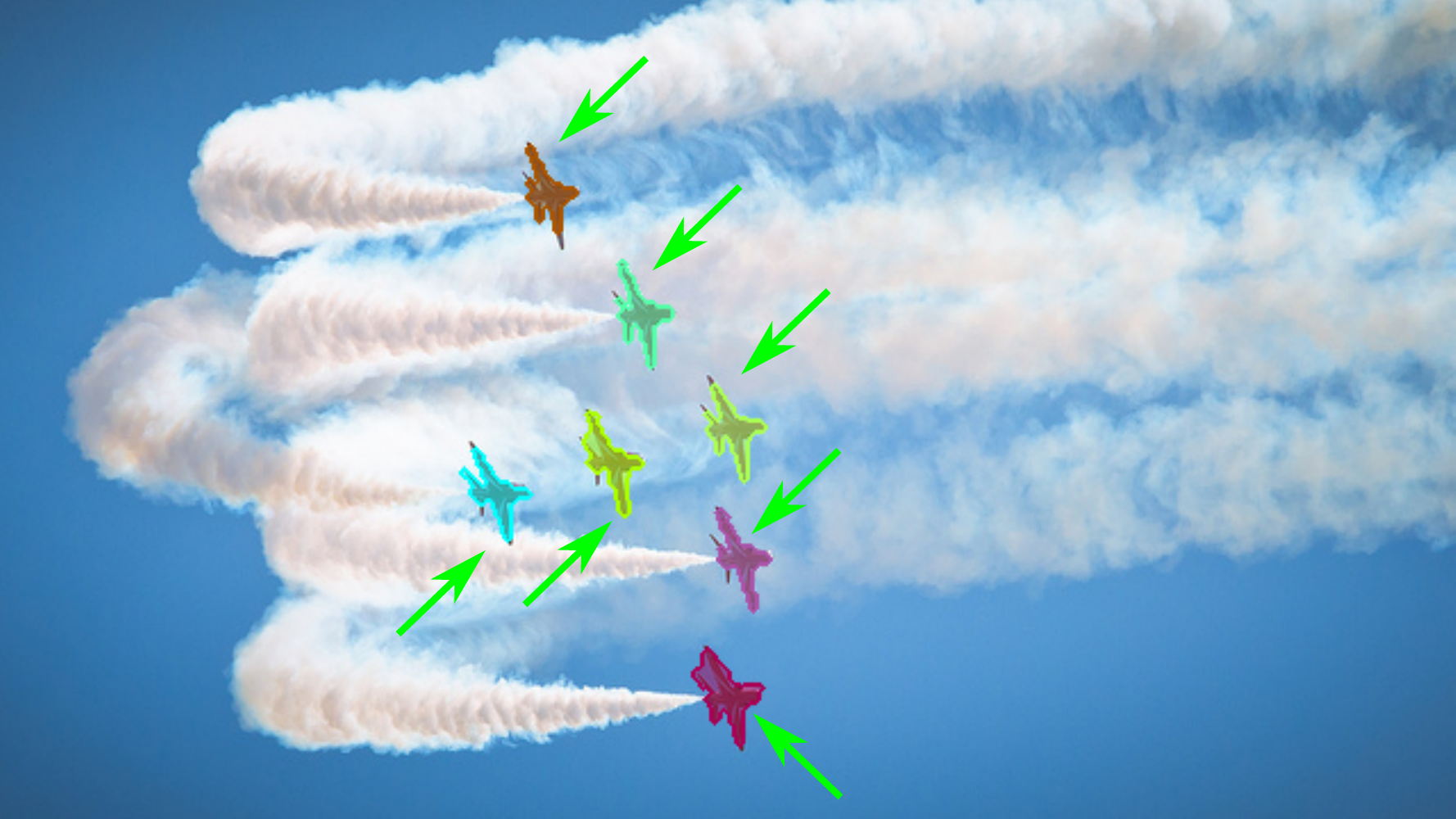} &
\includegraphics[width = .175\linewidth]{./figs/300659_gt}
\vspace{-1mm} \\ 
{\scriptsize FastMask} &{\scriptsize AttentionMask}&{\scriptsize Ours with FH}&{\scriptsize Ours with DeepFH}&{\scriptsize Ground Truth}
\end{tabular}
\caption{Qualitative results of FastMask~\cite{Hu2017-fastmask}, AttentionMask~\cite{WilmsFrintropACCV2018} and our approach with FH~\cite{felzenszwalb2004efficient} and our new DeepFH superpixels on the COCO dataset with LVIS annotations. The filled colored contours denote found objects, while not filled red contours denote missed objects. The green arrows highlight locations showing the strength of our proposals adhering well to object boundaries, while the red arrows denote the same locations in the results of selected other systems.
}
\label{fig:qualRes}
\end{figure*}

Investigating the proposal's boundary adherence, we generate a segmentation per found object for each of the systems using the best proposal per found object. We compare these segmentations to the segmentations generated from the ground truth objects. The results of this experiment (see Tab.~\ref{tab:resultsSeg}) allow to analyze the proposal masks using typical segmentation measures like BR and UE. The numbers show that both of our proposed systems outperform all other DL-based systems by a large margin (e.g., $+0.153$ in BR compared to AttentionMask). Thus, our proposals fit the ground truth objects better in terms of BR and UE. Only the not DL-based method MCG generates similar results in terms of BR, however, missing a larger amount of objects (see Tab.~\ref{tab:resultsLVIS}). Comparing our refinement using FH and DeepFH segmentations shows that the DeepFH segmentations further improve BR ($+3\%$) and UE ($-3\%$). Thus, our refined proposals based on DeepFH adhere better to object boundaries compared to all other methods.

\begin{table}[t]
\renewcommand{\arraystretch}{1.2}
\centering
\caption{Comparison of the best proposals per image as segmentations using segmentation measures boundary recall (BR) and undersegmentation error (UE) on the LVIS dataset.  Best and second-best results marked with \textbf{bold font} and \textit{italic font} respectively.}
\label{tab:resultsSeg}
{
\begin{tabular}{@{}lcccc@{}}
\hline
Method &  Backbone & BR$\uparrow$ & UE$\downarrow$ \\
\hline
MCG~\cite{pont2017multiscale} & - & \textit{0.685}  & 0.073 \\
\hline
DeepMask~\cite{Pinheiro2015-deepmask} & ResNet-50 & 0.488 & 0.087 \\
SharpMask~\cite{Pinheiro2016-sharpmask} & ResNet-50 & 0.561 & 0.080 \\
FastMask~\cite{Hu2017-fastmask} & ResNet-50 & 0.510  & 0.084 \\
AttentionMask~\cite{WilmsFrintropACCV2018} & ResNet-50  & 0.568 &  0.070  \\
\hline
AttentionMask~\cite{WilmsFrintropACCV2018} & ResNet-34  & 0.547 &  0.075 \\
Ours with FH~\cite{felzenszwalb2004efficient}& ResNet-34 &  0.681  & \textit{0.068}  \\
Ours with DeepFH& ResNet-34 &  \textbf{0.700}  & \textbf{0.066}  \\
\hline
\end{tabular}
}
\end{table}

The qualitative results in Fig.~\ref{fig:qualRes} support these findings. It is clearly visible that our proposed refinement improves the boundary adherence compared to other systems. In the first row, e.g., the tennis player's limbs are only precisely captured using our refinement with DeepFH. Similarly, in the second row the giraffes' heads and limbs are captured in more detail using our refinement with DeepFH compared to the imprecise segmentations of FastMask and AttentionMask. Even the small ears and most of the ossicones are precisely captured. Compared to the refinement with FH, less superpixels were misclassified. Similar effects are visible along the contour of the bus in the third row, which is not segmented as a blob like in the AttentionMask and FastMask results. Focusing on smaller objects, the final two rows show the precise segmentation of a jumping dog even capturing its thin tail and the detailed segmentations of tiny airplanes using our refinement. Therefore, as the results in Tab.~\ref{tab:resultsLVIS} and Tab.~\ref{tab:resultsSeg} indicated, our proposed refinement with DeepFH captures objects of all scales in more detail than other approaches.

\subsection{Ablation Studies}
\label{sec:eval_lvis_abl}
After discussing the main results, we present ablation studies regarding the segmentation and the post-processing of our system. Other studies regarding design choices were already presented in Sec.~\ref{sec:meth_deepfh} and Sec.~\ref{sec:meth_refinement}. All studies, except for the analysis of the post-processing, were carried out on our COCO validation set using LVIS annotations.

\subsubsection{Segmentation Algorithms for Refinement}
\label{sec:eval_lvis_abl_seg}
In this study, we evaluate the use of different segmentation algorithms and numbers of superpixels for the object proposal refinement based on AR. First, we compare our DeepFH-based refinement system with different numbers of superpixels. The results for 8000 to 500 superpixels and 4000 to 250 superpixels are presented in the first part of Tab.~\ref{tab:ablInflSeg}. It is visible that more superpixels have a positive influence on the results. However, more than 8000 superpixels are impossible due to memory limitations. In a second experiment, we add all ground truth edges to both variations. The second part of Tab.~\ref{tab:ablInflSeg} shows the results. Interestingly, a smaller number of superpixels generates significantly better results. Thus, a strong oversegmentation is not desirable for our proposed refinement system.

As a third step, we evaluate other segmentation methods presented in Sec.~\ref{sec:relWork_superpixels} with our refinement system. The results are shown in the third and fourth part of Tab.~\ref{tab:ablInflSeg}. The segmentation methods in the fourth part do not utilize deep features. The results show that the methods ETPS, ERS, SLIC and SEEDS generating rather compact superpixels of uniform size perform worse than original FH. Our DeepFH outperforms all those methods by utilizing deep features. Comparing DeepFH to other DL-based superpixel approaches like SSN, SEAL and SuperFCN, the results also show a better performance of DeepFH in object proposal refinement. This is in line with the non DL-based results, since SSN, SEAL and SuperFCN generate rather compact superpixels of uniform size. Thus, for the proposed refinement system a FH-based segmentation and specifically DeepFH are better suited than other segmentation methods.

\begin{table}[t]
\renewcommand{\arraystretch}{1.2}
\centering
\caption{Results of different segmentation methods and numbers of superpixels used in our refinement approach. Best result without and with ground truth are marked with \textbf{bold font} and \textit{italic font} respectively}
\label{tab:ablInflSeg}
{
\begin{tabular}{@{}lcccc@{}}
\hline
Segmentation & \#Superpixels & AR@100$\uparrow$  & AR@1000$\uparrow$ \\ 
\hline
DeepFH & $8000-500$  &  \textbf{0.223} & \textbf{0.330} \\
DeepFH & $4000-250$ & 0.219 & 0.322 \\
\hline 
DeepFH with GT & $8000-500$ &  0.244 & 0.361 \\
DeepFH with GT & $4000-250$  &  \textit{0.272} & \textit{0.405} \\
\hline
SpxFCN~\cite{yang2020superpixel} & $8000-500$ &   0.207 & 0.302 \\
SSN~\cite{jampani2018superpixel} & $8000-500$ & 0.220 & 0.319 \\
SEAL~\cite{tu2018learning} & $8000-500$ &   0.215 & 0.310 \\
\hline
FH~\cite{felzenszwalb2004efficient} & $8000-500$ &  0.217 & 0.318 \\
ETPS~\cite{yao2015real} & $8000-500$ & 0.214 & 0.311 \\
ERS~\cite{liu2011entropy} & $8000-500$  &  0.205 & 0.302 \\
SEEDS~\cite{van2012seeds} & $8000-500$ &  0.200 & 0.284 \\
SLIC~\cite{achanta2012slic} & $8000-500$ &  0.196 & 0.292 \\
\hline
\end{tabular}
}
\end{table}

\label{sec:eval_lvis_abl_segQuality}
\begin{figure*}[t]
\centering
\begin{subfigure}{0.24\textwidth}
\centering
\includegraphics[width=\textwidth]{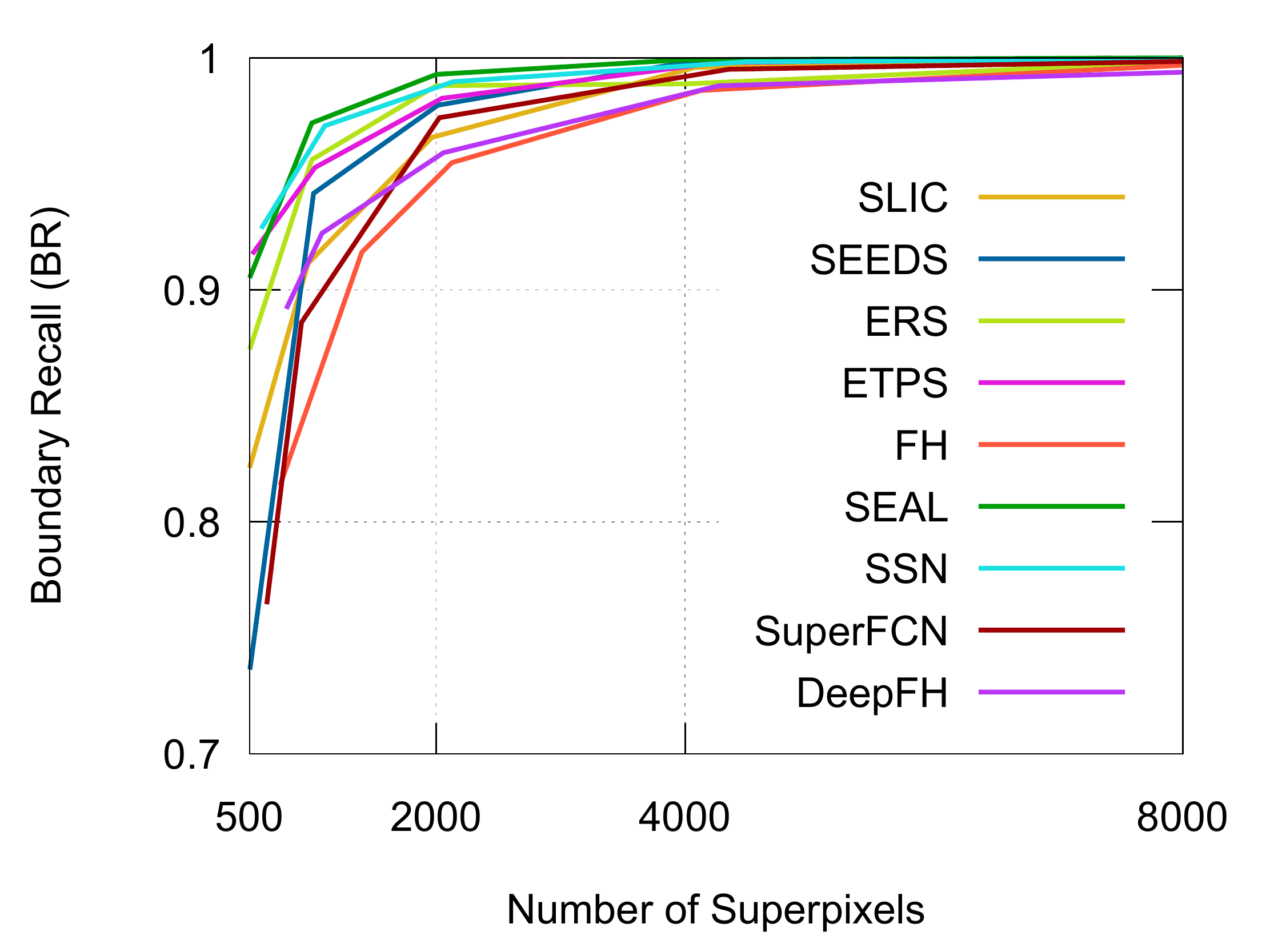}
\caption{}
\label{fig:plotRec}
\end{subfigure}
\hfil
\begin{subfigure}{0.24\textwidth}
\centering
\includegraphics[width=\textwidth]{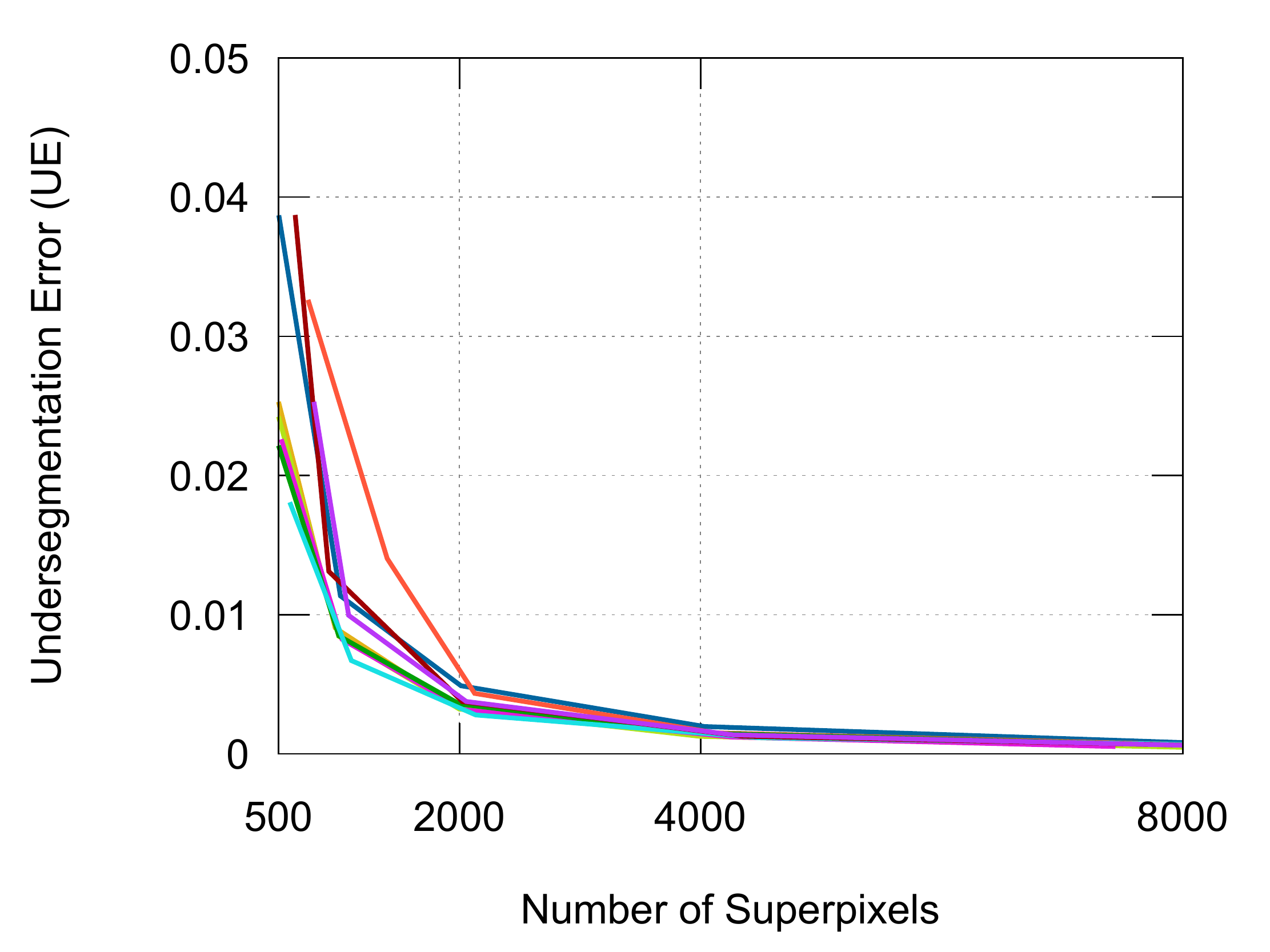}
\caption{}
\label{fig:plotUE}
\end{subfigure}
\hfil
\begin{subfigure}{0.24\textwidth}
\centering
\includegraphics[width=\textwidth]{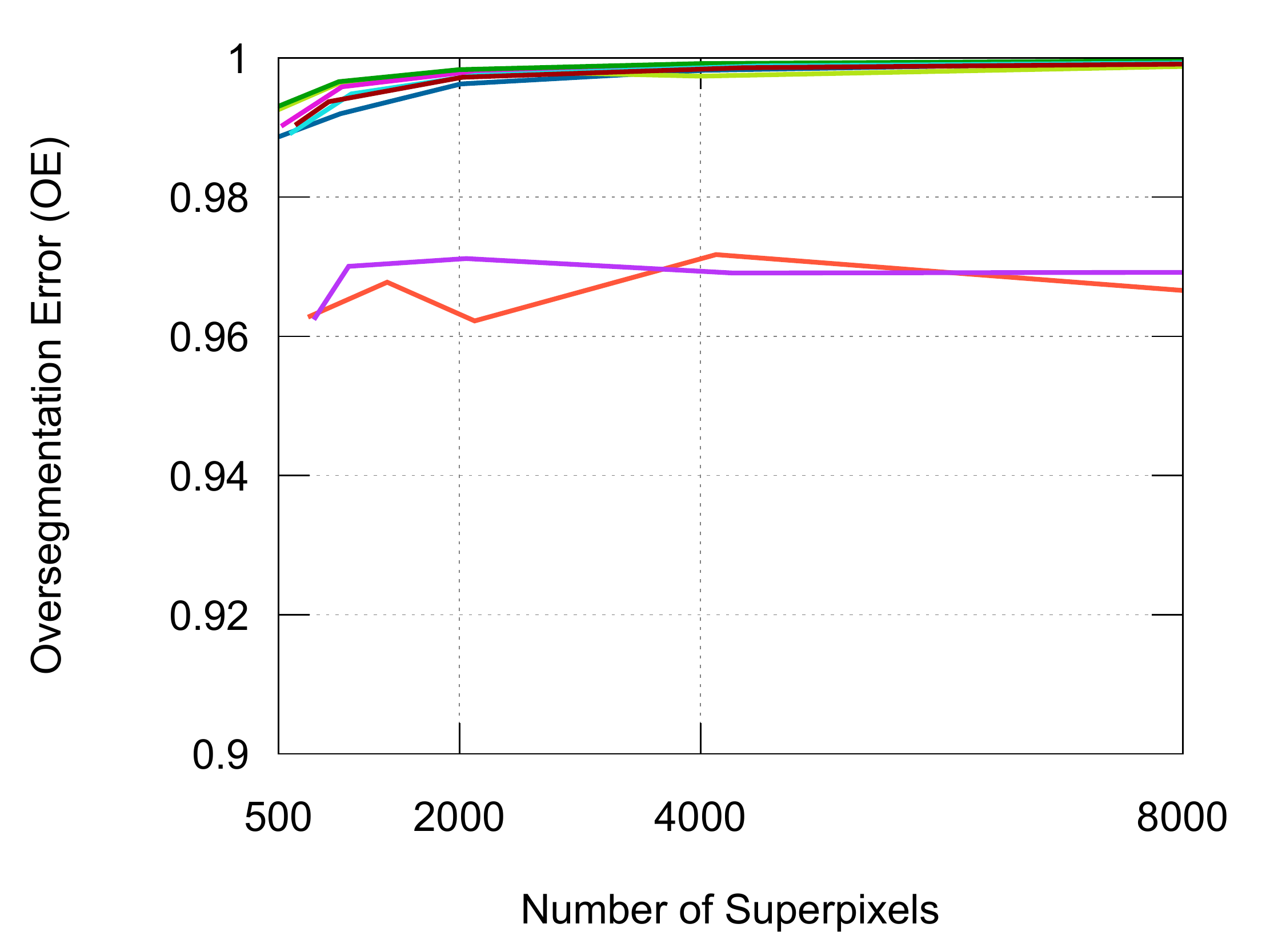}
\caption{}
\label{fig:plotOE}
\end{subfigure}
\hfil
\begin{subfigure}{0.24\textwidth}
\centering
\includegraphics[width=\textwidth]{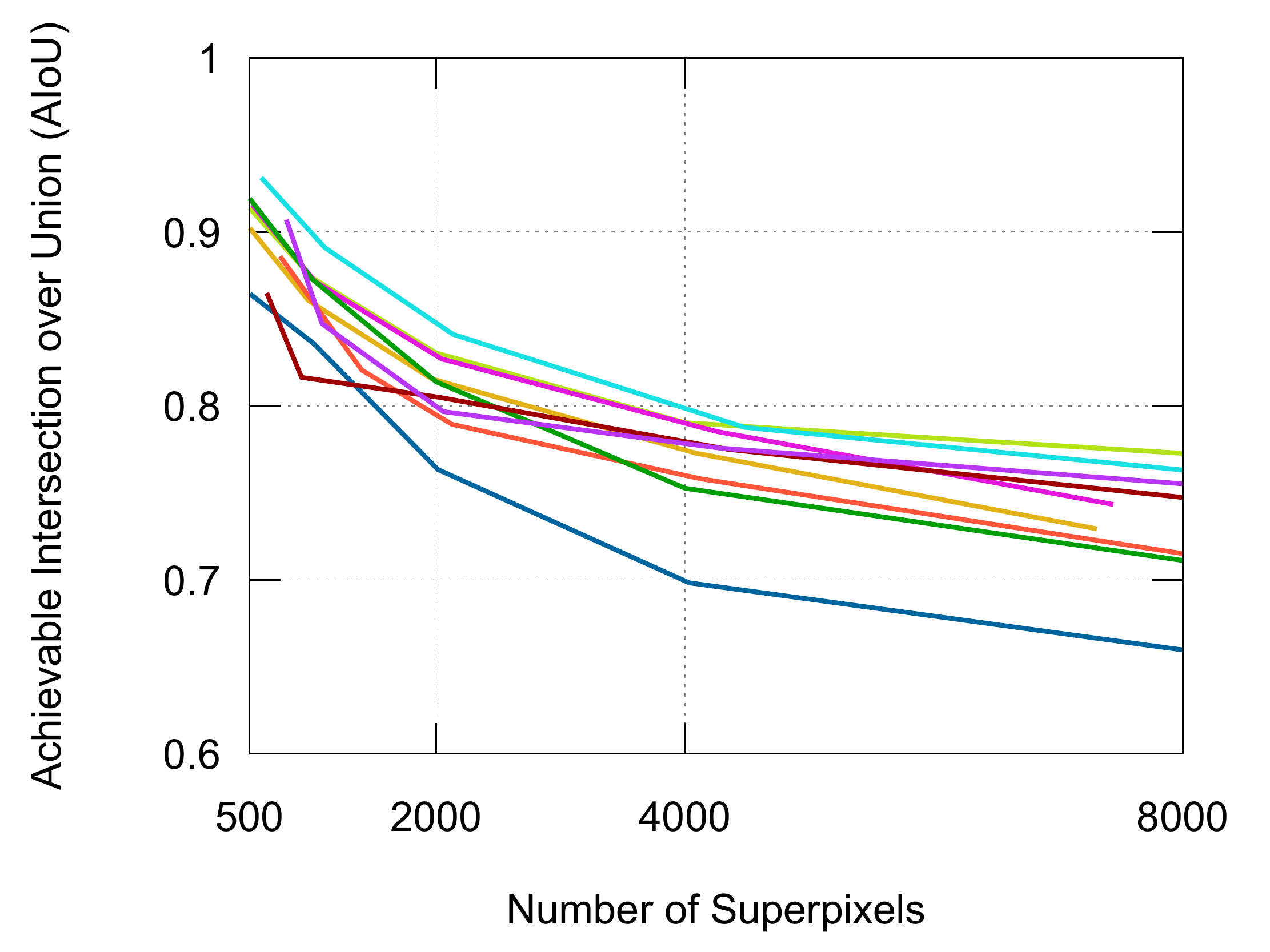}
\caption{}
\label{fig:plotAIoU}
\end{subfigure}
\caption{Results of different segmentation algorithms on the LVIS annotations for varying numbers of superpixels in terms of BR, UE, OE and AIoU.}
\label{fig:plotsSeg}
\end{figure*}

\begin{figure*}[t]
\centering
\begin{tabular}{cccc}
\includegraphics[width = .22\linewidth]{./figs/155571} &
\includegraphics[width = .22\linewidth]{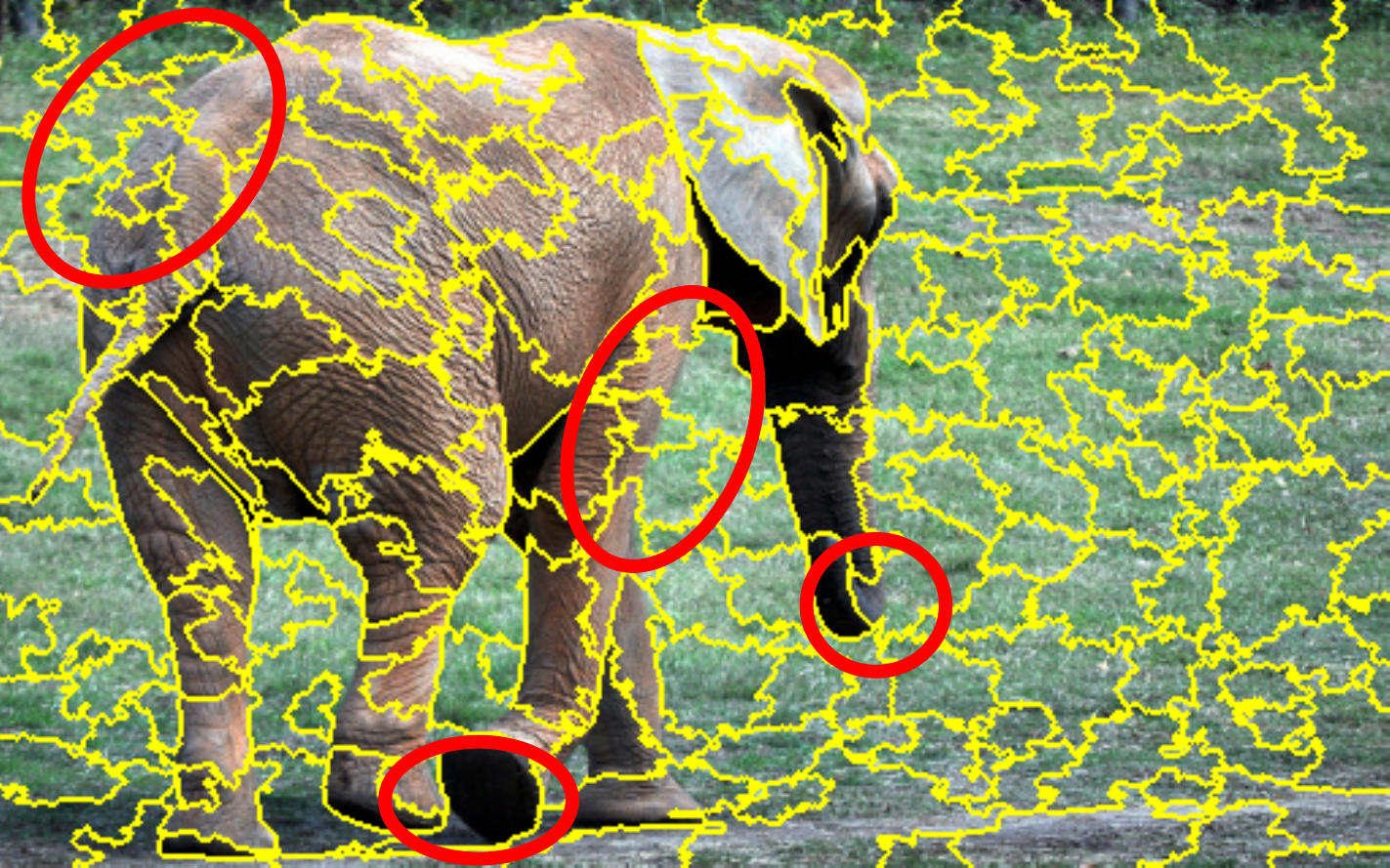} &
\includegraphics[width = .22\linewidth]{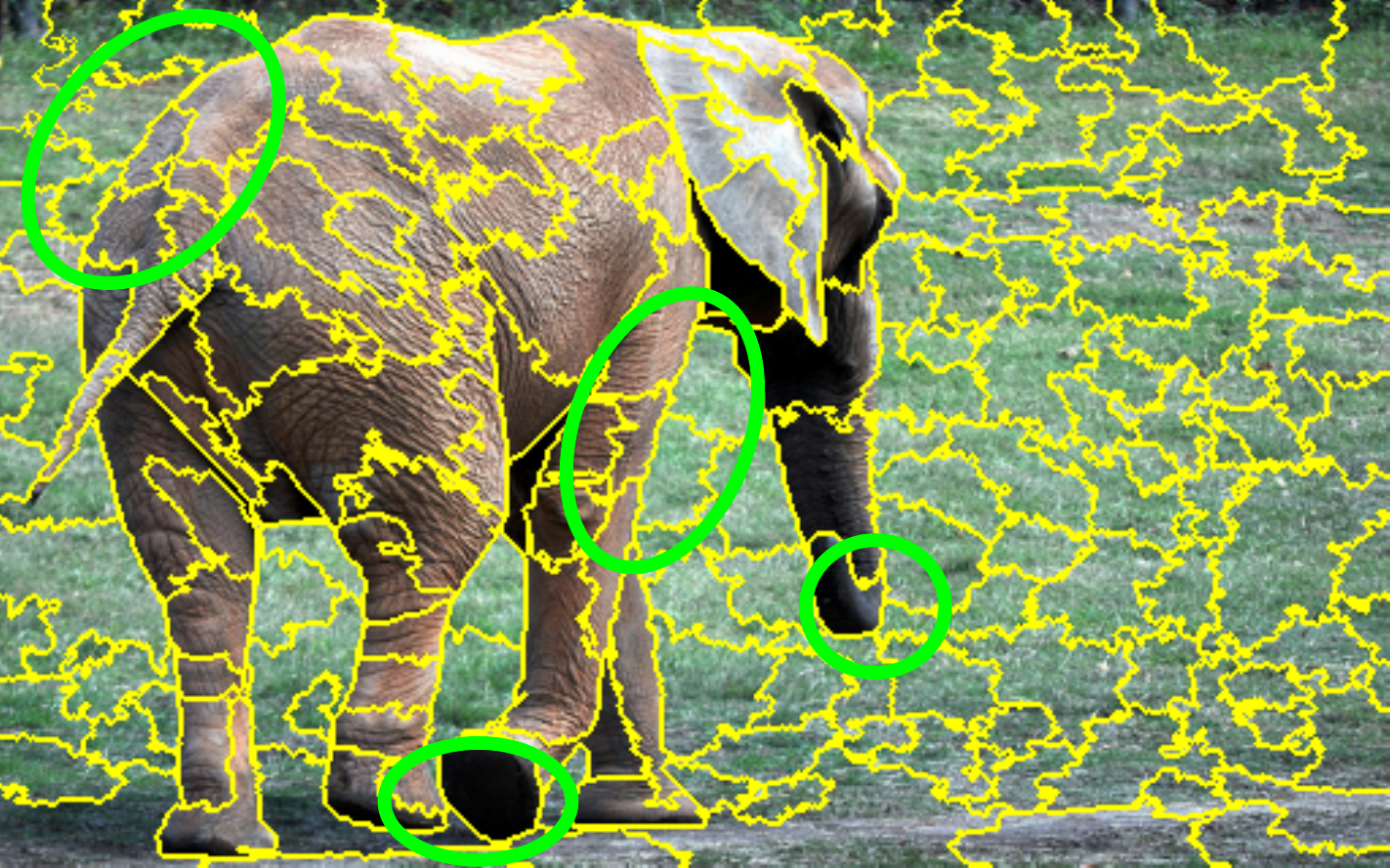} & 
\includegraphics[width = .22\linewidth]{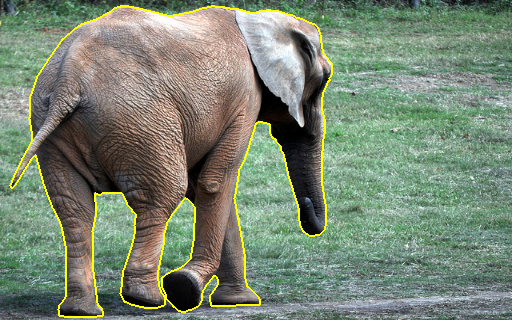}
\\
\includegraphics[width = .22\linewidth]{./figs/469652} &
\includegraphics[width = .22\linewidth]{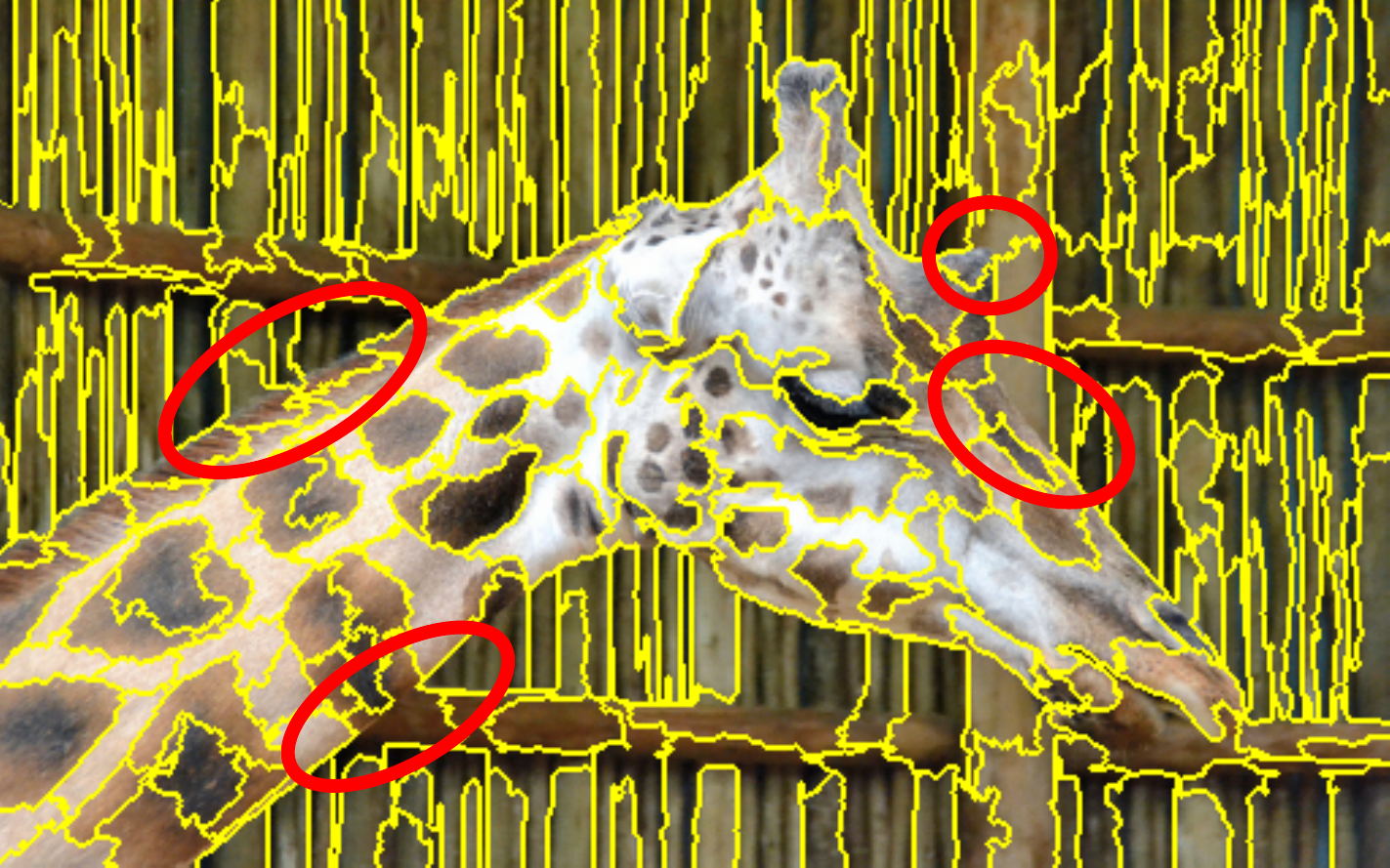} &
\includegraphics[width = .22\linewidth]{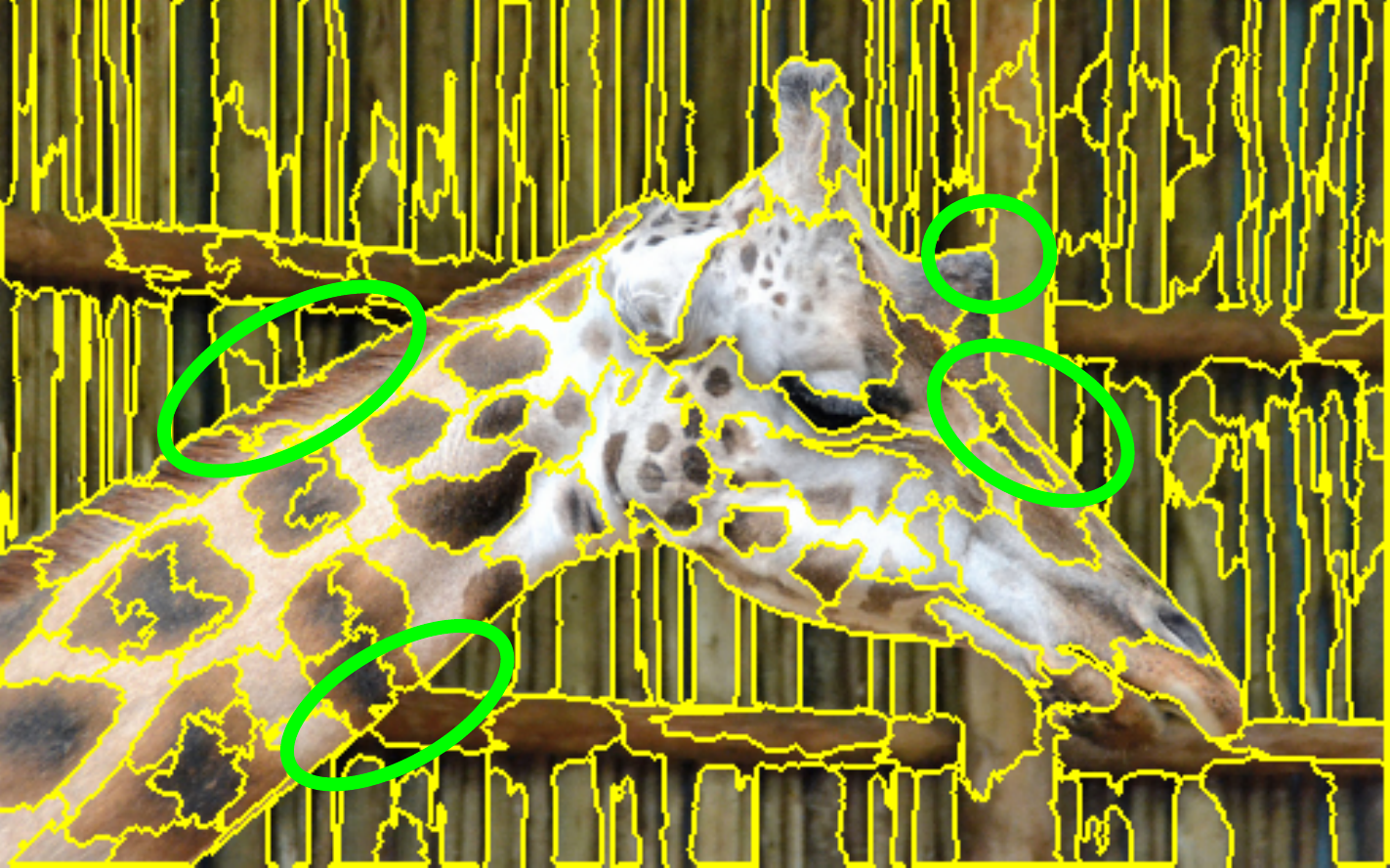} &
\includegraphics[width = .22\linewidth]{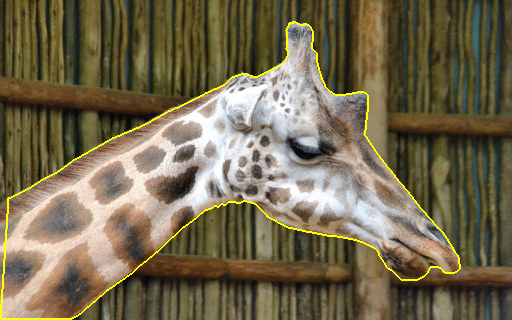} 
\vspace{-1mm} \\ 
{\scriptsize Image} &{\scriptsize FH}&{\scriptsize DeepFH}&{\scriptsize Ground Truth}
\end{tabular}
\caption{Qualitative results of FH~\cite{felzenszwalb2004efficient} and our proposed DeepFH segmentation on COCO images with LVIS annotations. Red circles denote segmentation errors in FH, while green circles denote the same location in the DeepFH results.
}
\label{fig:qualResSeg}
\end{figure*}

\subsubsection{Segmentation Quality}
After showing the superiority of DeepFH segmentations in object proposal refinement, we evaluate the different segmentations in terms of segmentation measures BR, UE and OE on the LVIS annotations. Ground truth segmentations represent the joint segmentations of the annotated objects. The results are presented in Fig.~\ref{fig:plotsSeg}. SSN and SEAL perform best in terms of BR (see Fig.~\ref{fig:plotRec}) and UE (see Fig.~\ref{fig:plotUE}). This is in line with the results of~\cite{jampani2018superpixel,tu2018learning} on segmentation datasets. Thus, those segmentations adhere generally better to the object boundaries. Both, FH and DeepFH do not perform very well in terms of BR and UE. However, evaluating the strength of oversegmentation in terms of OE (see. Fig.~\ref{fig:plotOE}), FH and DeepFH outperform all other methods. Thus, due to the design of FH and DeepFH, both segmentations do not artificially oversegment uniform areas as strongly. This leads to a much smaller OE and to better results on the downstream task, since many superpixels covering larger parts of background reach the object boundary. Therefore, our superpixel refinement prefers segmentations with a smaller OE, despite a slightly worse segmentation accuracy in terms of BR and UE.

Evaluating a segmentation's suitability for super\-pixel-based refinement of object proposals, we proposed AIoU as an upper limit for the average IoU between proposals and annotations. The results in terms of AIoU are presented in Fig.~\ref{fig:plotAIoU}. Again, SSN performs very well, while FH and DeepFH generate mediocre results. This is also supported by the results in Tab.~\ref{tab:ablAIoU}, which show the average AIoU across the scales for selected segmentations. Tab.~\ref{tab:ablAIoU} also shows the \mbox{AVGIoU}, the average IoU between an annotation and the best fitting proposal. In contrast to the AIoU, these numbers favor DeepFH and FH segmentations. Thus, those segmentations realize a larger portion of their potential (AIoU), leading to a larger efficiency in terms of $\text{AVGIoU}/\text{AIoU}$. Still, it is clearly visible that all segmentations realize only slightly above $50\%$ of their AIoU.

\begin{table}[t]
\renewcommand{\arraystretch}{1.2}
\centering
\caption{Results of selected segmentations with the proposed refinement approach on top of AttentionMask. Evaluation in terms of average IoU (AVGIoU) between a ground truth object and the best fitting proposal, the achievable IoU for a given segmentation (AIoU) as well as the efficiency $\text{AVGIoU}/\text{AIoU}$. Best and second-best results marked with \textbf{bold font} and \textit{italic font} respectively.}
\label{tab:ablAIoU}
{
\begin{tabular}{@{}lccc@{}}
\hline
Segmentation & AVGIoU$\uparrow$ & AIoU$\uparrow$ & Efficiency$\uparrow$ \\ 
\hline
DeepFH & \textbf{0.448} &  0.790 &\textit{ 0.568} \\
\hline
SSN~\cite{jampani2018superpixel} & \textit{0.443} &  \textit{0.809} & 0.547 \\
SEAL~\cite{tu2018learning} & 0.438 &  0.777 & 0.563 \\
\hline
FH~\cite{felzenszwalb2004efficient} & 0.441 &  0.766 & \textbf{0.577} \\
ETPS~\cite{yao2015real} & 0.424 &  0.799 & 0.530 \\
ERS~\cite{liu2011entropy} & 0.436 &  \textbf{0.811} & 0.538 \\
\hline
\end{tabular}
}
\end{table}

\subsubsection{FH vs. DeepFH Segmentations}
\label{sec:eval_lvis_abl_fh_deepFH}
We already quantitatively evaluated the difference between our proposed DeepFH and the original FH in terms of BR, UE, OE and AIoU on the LVIS annotations. Most of these measures showed an improved performance for DeepFH compared to FH. Highlighting this improvement, Fig.\ref{fig:qualResSeg} shows qualitative segmentation results on the COCO dataset with LVIS annotations. In both images, it is visible that the FH segmentation results in segmentation errors along the object's contour. For instance, in the first image along the back of the elephant, multiple superpixels cover both object and background. This can be attributed to the low contrast between object and background. Similarly, the bottom of the elephant's lifted foot is not segmented properly. Those effects are fixed in the DeepFH segmentation. The image in the second row shows a highly textured environment, where object and background share similar colors. The FH segmentation reflects this with several segmentation errors along the giraffes' neck and head. DeepFH properly distinguishes between object and background by utilizing semantics in deep features.

\subsubsection{Post-Processing}
\label{sec:eval_lvis_abl_pp}
In Sec.~\ref{sec:meth_refinement_combination}, multiple post-processing steps on top of the refinement system were introduced. Tab.~\ref{tab:ablPP} presents the detailed results of each post-processing step on the final result. First, the bilateral filtering on superpixel-level improves the results significantly, as the numbers indicate. This step mainly adjusts the results for adjacent superpixels with similar color. The next two steps apply opening and closing to the proposals eliminating segmentation artifacts. This has a rather large effect in qualitative results, however, the quantitative results also show an improvement. The final NMS for near duplicate removal reduces the AR@1000 slightly as expected, however, the more relevant AR@10 and AR@100 significantly improve. Therefore, Tab.~\ref{tab:ablPP} shows improvements for all individual post-processing steps.

\begin{table}[t]
\renewcommand{\arraystretch}{1.2}
\centering
\caption{Results of the proposed refinement approach using different post-processing steps on the test dataset. Best result marked with \textbf{bold font}.
}
\label{tab:ablPP}
{
\begin{tabular}{@{}lcccccc@{}}
\hline
Post-processing & AR@10$\uparrow$ & AR@100$\uparrow$  & AR@1000$\uparrow$ \\
\hline
none & 0.076  & 0.189  & 0.288 \\
bilateral filtering & 0.079  & 0.196  & 0.304 \\
+ opening & 0.080 &  0.197 & 0.306 \\
+ closing & 0.080 &  0.199 & \textbf{0.307} \\
+ NMS & \textbf{0.094} & \textbf{0.213} & 0.304 \\
\hline
\end{tabular}
}
\end{table}

\section{Conclusion}
\label{sec:conclusion}
In this paper, we tackle the problem of class-agnostic object proposal generation as a typical first step in object detection pipelines. We introduced the problem of imprecise object proposals due to subsampling in CNN-based systems and showed that typical solutions are not feasible with thousands of proposals. Therefore, we propose a superpixel-based refinement approach on top of the AttentionMask system. The refinement utilizes our novel DeepFH segmentation and superpixel pooling on deep features. Thereby, we combine coarse initial proposals and precise superpixel segmentations to generate precise object proposals.

The results on the COCO dataset with LVIS annotations showed the improvements of our overall system compared to state-of-the-art object proposal systems. We further showed the increased segmentation quality of the proposals with typical segmentation measures. Finally, we evaluated the influence of the chosen segmentation method in detail. For best performance, a segmentation with less artificial oversegmentation of uniform areas is necessary, like FH and our proposed DeepFH. In addition, we showed an improved performance of our DeepFH segmentation compared to FH for the object proposal refinement task and the general superpixel segmentation task. However, the results also indicated that just above $50\%$ of the segmentation's potential is utilized for the object proposal refinement. Thus, a more sophisticated superpixel classification is needed to reach the superpixels' potential on this task.


\bibliographystyle{cas-model2-names}

\bibliography{lit.bib}

\end{document}